\def\year{2021}\relax
\documentclass[letterpaper]{article} 
\usepackage{aaai21}  
\usepackage{times}  
\usepackage{helvet} 
\usepackage{courier}  
\usepackage[hyphens]{url}  
\usepackage{graphicx} 
\usepackage{natbib}  
\usepackage{caption} 
\usepackage{amsmath,amsthm,amssymb}
\usepackage{color}

\usepackage{array,varwidth,lipsum}

\usepackage[linesnumbered,ruled,vlined]{algorithm2e}
\usepackage[switch]{lineno}

\newtheorem{theorem}{Theorem}
\newtheorem{lemma}{Lemma}

\newtheorem{remark}{Remark}
\usepackage{amsthm}
\usepackage{graphicx}
\usepackage{subfig}

\usepackage{amsmath}
\usepackage{amssymb}
\usepackage[utf8]{inputenc} 
\usepackage[T1]{fontenc}    
\usepackage{hyperref}       
\usepackage{cleveref}
\usepackage{url}            
\usepackage{booktabs}       
\usepackage{amsfonts}       
\usepackage{nicefrac}       
\DeclareMathOperator{\Tr}{Tr}
\usepackage{color}
\usepackage{microtype}      

\newtheorem{result}{Result}

\newtheorem{assumption}{Assumption}

\usepackage[linesnumbered,ruled,vlined]{algorithm2e}
\newcommand{\HB}{\mathbf{H}}
\newcommand{\RB}{\mathbb{R}}
\newcommand{\EB}{\mathbb{E}}
\newcommand{\DB}{\mathbf{D}}
\newcommand{\IB}{\mathbf{I}}
\newcommand{\AB}{\mathbf{A}}
\newcommand{\WB}{\mathbf{W}}
\newcommand{\gb}{\mathbf{g}}

\newcommand{\eb}{\mathbf{e}}

\newcommand{\ab}{\mathbf{a}}
\newcommand{\bb}{\mathbf{b}}
\newcommand{\cb}{\mathbf{c}}
\newcommand{\yb}{\mathbf{y}}
\newcommand{\vb}{\mathbf{v}}
\newcommand{\GB}{\mathbf{G}}
\newcommand{\UB}{\mathbf{U}}

\newcommand{\0}{\mathbf{0}}
\DeclareMathOperator{\E}{\mathbb{E}}
\newcommand{\EC}{\mathcal{E}}
\newcommand{\NC}{\mathcal{N}}
\newcommand{\WC}{\mathcal{W}}
\newcommand{\VC}{\mathcal{V}}

\newcommand\norm[1]{\left\lVert#1\right\rVert_2}

\newcommand\normpsi[1]{\left\lVert#1\right\rVert_{\psi_2}}

\newcommand\normpsia[1]{\left\lVert#1\right\rVert_{\psi_1}}

\newcommand{\nn}{\nonumber}
\title{Learning Graph Neural Networks with Approximate Gradient Descent}
\author {
        Qunwei Li,\textsuperscript{\rm 1}
        Shaofeng Zou, \textsuperscript{\rm 2}
        Wenliang Zhong \textsuperscript{\rm 1} \\
}
\affiliations {
    \textsuperscript{\rm 1} Ant Group \\
    \textsuperscript{\rm 2}  University at Buffalo, the State University of New York \\
    qunwei.qw@antgroup.com, szou3@buffalo.edu, yice.zwl@antgroup.com
}
\begin{document}

\maketitle

\begin{abstract}
The first provably efficient algorithm for learning graph neural networks (GNNs) with one hidden layer for node information convolution is provided in this paper. Two types of GNNs are investigated, depending on whether labels are attached to nodes or graphs. A comprehensive framework for designing and analyzing convergence of GNN training algorithms is developed. The algorithm proposed is applicable to a wide range of activation functions including ReLU, Leaky ReLU, Sigmod, Softplus and Swish. It is shown that the proposed algorithm guarantees a linear convergence rate to the underlying true parameters of GNNs. For both types of GNNs, sample complexity in terms of the number of nodes or the number of graphs is characterized. The impact of feature dimension and GNN structure on the convergence rate is also theoretically characterized. Numerical experiments are further provided to validate our theoretical analysis.
\end{abstract}

\section{Introduction}
\label{sec:intro}
Recent success of deep neural network (DNN) has dramatically boosted its application in various application domains, e.g., computer vision~\cite{NIPS2012_4824}, natural language processing~\cite{young2018recent} and strategic reasoning with reinforcement learning \cite{Sutton2018,silver2017mastering}. Although DNN usually has millions of parameters and a highly non-convex objective function, simple optimization techniques, e.g., stochastic gradient descent and its variants, can efficiently train such a complicated structure successfully. However, there are many applications, where data is represented in the form of graphs. For example, in social networks, a graph-based learning system can exploit the interconnection between the users to make highly accurate inferences. DNN, when applied to graph based structural data, has to face with some new challenges, and cannot be directly applied. To address the new challenges, graph neural network (GNN) has been proposed recently and has been demonstrated to have massive expressive power on graphs. GNNs have been shown to be successful in many application domains, e.g., social networks, knowledge graph, recommender systems, molecular fingerprint, and protein interface networks \cite{wu2019comprehensive, xu2018powerful, zhou2018graph}. 



Although being successful in practical applications, theoretical understanding of GNNs is still lacking in the literature. There is a strong need to theoretically understand why GNNs work well on practical graph-structured data, and to further design provably efficient learning algorithms for GNN. 

However, even for commonly used DNN architectures, it is in general challenging to develop theoretical understanding of the learning dynamics and to learn simple neural networks \cite{pmlr-v80-goel18a}. Specifically, if we assume that the data is generated according to a teacher neural network, the goal is then to recover the true parameters of the teacher neural network. The major challenge lies in the highly non-convex nature of DNN's objective function. For example, in the agnostic setting with bounded distributions and square loss, learning a single ReLU with unit norm hidden weight vectors is notoriously difficult~\cite{pmlr-v65-goel17a}. It was proved by Brutzkus and Globerson~\cite{gautier2016globally} that the problem of distribution-free recover of the unknown weight vector for learning one hidden layer convolutional neural network (CNN)  is NP-hard, even if non-overlapping patch structure is assumed. At this point, a major open problem is still to understand mildest conditions and design of provably efficient algorithms that lead to learnability of neural networks in polynomial time. 


In this paper, we focus on the parameter recovery problem for learning GNNs where the training data is generated according to a teacher GNN. We design provably efficient algorithms to recover the true parameters of the teacher GNN. Our focus is to develop a comprehensive framework for designing and analyzing convergence of GNN training algorithms. 

\subsection{Main Contributions}
Our first contribution is the design and convergence analysis of an approximate gradient descent algorithm for training GNNs. The algorithm is based on the idea of inexact optimization and approximate training~\cite{schmidt2011convergence,qunwei2017convergence,cao2019tight}, and the major advantage is that it reduces computational complexity while guaranteeing convergence and learnability at the same time. We prove that the proposed algorithm recovers the underlying true parameters of the teacher network with a linear convergence rate up to statistical precision. The assumptions are mild, and can be easily satisfied in practice. Specifically, our analysis is applicable to a wide range of activation functions (see Assumptions \ref{assump:1} and \ref{assump:2}), e.g., ReLU, Leaky ReLU, Sigmod, Softplus and Swish. The analysis only requires that the activation functions to be monotonic increasing, and Lipschitz, and does not depend on the specific gradient computation involved in the activation function \cite{brutzkus2017globally,du2017convolutional,safran2018spurious}. 

Our second contribution is the introduction and the extension of the technique of approximate calculation in gradients for the algorithm in order to analyze GNNs. A similar idea was first proposed in~\cite{cao2019tight} to analyze the learnability of CNNs with  non-overlapping convolution patches. We highlight that the non-overlapping convolution process is very different from the nature that the feature convolution at nodes in GNNs is intrinsically overlapping, as nodes may share common neighbors.  The analysis framework in~\cite{cao2019tight} cannot be directly applied in GNN analysis. We extend the scope of the methodology and propose provably efficient algorithms to learn and analyze GNNs.

Our third contribution is that we investigate the empirical version of the problem where the estimation of parameters is based on $n$ independent samples. We provide uniform convergence results of the proposed algorithm with respect to the sample complexity. We also provide the parameter training dynamics with the proposed algorithm, and show that the training is provably stable.


    


To the best of the authors' knowledge, these theoretical results are the first sharp analyses of statistical efficiency of GNNs. For ease of presentation, we hereby informally present the main theorem in the paper as follows. We refer readers to Theorem~\ref{convergence_thm} for the precise statements.
\begin{theorem}[Main Theorem (Informal)] GNN is stably learnable with the proposed algorithms in linear time.
\end{theorem}

\subsection{Related Work}
There is a recent surge of interest in theoretically understanding properties of DNNs, e.g., hardness of learning~\cite{pmlr-v65-goel17a,song2017complexity}, landscape of neural networks~\cite{kawaguchi2016deep,choromanska2015loss,hardt2016identity,haeffele2015global,freeman2019topology,safran2016quality,zhou2017landscape,nguyen2017loss,nguyen2017lossb,ge2017learning,safran2018spurious,du2018power}, training dynamics using gradient descent approaches~\cite{tian2017analytical,zhong2017recovery,li2017convergence,du2018many}, and design of provable learning algorithms~\cite{goel2017eigenvalue,goel2017learning,zhang2015learning}. For non-convex optimization problems that satisfy strict saddle property, it was shown in \cite{du2017convolutional} and \cite{jin2017escape} that (stochastic) gradient descent converges in polynomial time. 
The landscape of neural networks were then extensively studied~\cite{soltanolkotabi2017learning,kawaguchi2016deep,choromanska2015loss,hardt2016identity,haeffele2015global,freeman2019topology,safran2016quality,zhou2017landscape,nguyen2017loss,nguyen2017lossb,ge2017learning,safran2018spurious,du2018power}. Specifically, algorithms were designed for specific neural network architectures, and their learning of a neural network in polynomial time and sample complexity were further characterized~\cite{pmlr-v65-goel17a,zhang2016l1,zhang2015learning,sedghi2014provable,janzamin2015beating,gautier2016globally,goel2017learning,du2017convolutional}.


However, to the best of our knowledge, there has been no related attempt to  understand the training dynamics and to design provably efficient learning algorithms for GNNs. In recent advances \cite{du2017convolutional,du2017gradient,pmlr-v80-goel18a,brutzkus2017globally,zhong2017learning}, the condition for (stochastic) gradient descent or its variants to recover the underlying true parameter under the teacher-student model in polynomial time were analyzed 
for CNNs. Specifically, for spherical Gaussian distribution and non-overlapping patch structures, gradient descent approach can recover the underlying true parameter in polynomial time for CNNs \cite{brutzkus2017globally,du2017gradient}. In \cite{cao2019tight}, the information theoretic limits and the computational complexity of learning a CNN were developed.  


Note that the GNN shares a similar convolutional structure as the CNN \cite{lecun1995convolutional}, and is therefore closely related to the CNN. Specifically, an image can be viewed as a graph grid where adjacent pixels are connected by edges. Similar to a 2D convolution for an image, graph convolution can also be performed on an image by taking an weighted average of information from adjacent pixels. Here, each pixel in an image can be viewed as a node in a graph, and its neighbors are ordered. The size of neighbors are then determined by the convolution filer size. However, different from image data, the neighbors of a node in a graph for GNNs are unordered, and vary in sizes. Moreover, information convolution in graph is by nature overlapping. Therefore, the analysis for CNNs is intrinsically different from and cannot be directly carried over to the analysis for GNNs. We will elaborate this in this paper.

\subsection{Notations}
Let $\norm{\cdot}$ denote the Euclidean norm of a finite-dimensional vector or a matrix. For a positive semidefinite matrix $X$, we use $\sigma_m(X)$ as its nonzero smallest eigenvalue 
 and $\sigma_M(X)$ as its nonzero largest eigenvalue in sequel. Throughout this paper, we use capital letters to denote matrices, lower case letters to denote vectors, and lower case Greek letters to denote scalars. We denote $x\wedge y \triangleq \min\{x,y\}$.

\section{The Graph Neural Network}
\label{sec:statement}
In this section, we formalize the GNN model with one layer of node information convolution.

\subsection{Node-level GNN (NGNN) with One Graph}
We first investigate one graph with $n$ nodes, and node $i$ has a feature $\HB_{i}\in \RB^d$ with dimension  $d$. We define $\HB \in \RB^{n\times d}$  as the node feature matrix of the graph. To learn a GNN, a node first collects all the features from its neighboring nodes, and updates its own feature with a shared local transition function among these nodes. This process is termed as graph convolution and can be expressed as
\begin{align}
	\hat \HB_i = \sigma \left(\frac{1}{\left| \NC_i \right| }\sum_{j\in \NC_i} \HB_{j} \WB\right),
\end{align} 
where $\sigma$ is the activation function, $\NC_i$ is the set containing node $i$ and all its neighboring nodes, and $\WB\in \RB^{d\times d_{out}}$ represents the local transition function which usually is a fully connected network taking in the averaged node feature and outputs the updated feature. After this process, the node feature gets updated with a change in dimension from $d$ to $d_{out}$. The above process of graph convolution can be written in a global point of view as
\begin{align}
\hat \HB = \sigma (\DB^{-1}\AB\HB \WB),
\end{align}
where
 $\DB\in \RB^{n\times n}$ is the degree matrix of the graph and $\AB\in \RB^{n\times n}$ is the corresponding adjacency matrix. Here, $\DB^{-1}\AB\HB$ is the operation that one node updates its feature by taking the average of its own features and features of its neighbors. 

After the graph convolution, each node has its own feature updated, by incorporating local structural information of the graph and features from its neighboring nodes. Then, the updated node feature is passed into a fully connected layer $\vb \in \RB^{d_{out}}$. Thus, the output of the entire graph neural network for node-level tasks is 
\begin{align}
    \hat \yb = \sigma (\DB^{-1}\AB\HB \WB) \vb \in \RB^n.
\end{align}

As we have $\emph{one}$ graph, suppose the graph has a node-level label vector $\yb\in \RB^n$, and we have $n$ ground truth data pairs $\{\HB_i, \yb_i\}_{i=1}^n$. We assume that the node feature matrix $\HB\in \RB^{n\times d}$ is generated
independently from the standard Gaussian distribution, and the corresponding output $\yb\in \RB^n$
is generated from the teacher network with true parameters $\WB^\ast$ and $\vb^\ast$ as follows
\begin{align}
\yb =  \sigma (\DB^{-1}\AB\HB \WB_\ast) \vb_\ast +\epsilon.
\end{align}
Here, $\{\epsilon_i\}_{i=1}^n$ are independent sub-Gaussian white noises with $\psi_2$ norm $\nu$, which is a broader class including Gaussian noise but far less restrictive. Without loss of generality, we assume that $\|\WB_\ast\|_2 =  1$ in this paper. 

\subsection{Graph-level GNN (GGNN) with Multiple Graphs}
We then investigate with $n$ graphs, and the $j$-th graph has $n_j$ nodes with the node feature matrix  $\HB_{j} \in \RB^{n_j\times d}$. 
Similar to the convolution for the case of one graph for NGNN, the $j$-th graph updates its node features by the following convolution process
\begin{align}
    \hat \HB_j = \sigma (\DB_j^{-1}\AB_j\HB_j \WB),
\end{align}
where $\sigma$ is the activation function, $\DB_j\in \RB^{n_j\times n_j}$ is the degree matrix of $j$-th graph, and $\AB_j \in \RB^{n_j\times n_j}$ is the corresponding adjacency matrix. Here, $\DB_j^{-1}\AB_j\HB_j$ is the operation that one node in the $j$-th graph updates its feature by taking the average of its own features and the features of its neighbors. $\WB\in \RB^{d\times d_{out}}$ represents the local transition function which is shared among different nodes across different graphs.

For GGNN, graph $j$ with $n_j$ node features has to aggregate the node features such that the graph has a unique graph-level representation. In the following, we discuss several aggregation methods which are most widely used in the literature.

\emph{\bf Particular node feature as graph embedding.} This method picks a specific node feature to represent the global graph embedding. Define $\gb_j\in R^{d}$ as the embedding for the $j$-th graph, which in this case is expressed as
\begin{align}
    \gb_j^T = \ab_{j} \sigma (\DB_j^{-1}\AB_j\HB_j \WB),
\end{align}
where $\ab_{j} \in R^{1\times n_j}$ is a row vector with ``1'' as its $i$-th element and otherwise ``0''. Consequently, the $i$-th node feature is picked to represent the $j$-th graph for follow-up graph-level tasks.


{\bf  Attention-based Aggregation.}
In this approach, the graph embedding is a weighted sum of the node features and the weight is usually termed as attention that is learned from the corresponding node feature. 
The attention-based weighted feature aggregation for the $j$-th graph can be expressed as
\begin{align}
    \gb_j^T = \ab_j \sigma (\DB_j^{-1}\AB_j\HB_j \WB),
\end{align}
where $\ab_j\in R_+^{1\times n_j}$ is an attention row vector for all the nodes, with non-negative elements summing up to 1.

{\bf Averaging.}
Averaging is a special case of the attention based aggregation where the attention is identical for all the node features. Thus, we can write
\begin{align}
    \gb_j^T = \frac{1}{n_j}\mathbf{1}_{n_j}\sigma (\DB_j^{-1}\AB_j\HB_j \WB),
\end{align}
where $\mathbf{1}_{n_j}$ is a $n_j$ dimensional row vector of all ones.

After the graph convolution and the node feature aggregation, each graph has its own embedding, incorporating its local structural information and all its node features. Then, the graph embedding is passed into a fully connected layer $\vb \in \RB^{d}$. Thus, the output of the $j$-th graph for graph-level tasks is 
\begin{align}
\hat y_j = \ab_j \sigma ( \DB_j^{-1}\AB_j\HB_j \WB) \vb \in \RB.
\end{align}

As we have $\emph{multiple}$ graphs, suppose each graph has a graph-level label $y_j \in \RB$, and we have $n$ ground truth data pairs $\{\HB_j, y_j\}_{j=1}^n$. We assume that each node feature matrix $\HB_j\in \RB^{n_j\times d}$ is generated
independently from the standard Gaussian distribution, and the corresponding output $y_j\in \RB$
is generated from the teacher network with true parameters $\WB^\ast$ and $\vb^\ast$ as follows,
\begin{align}
y_j =   \ab_j\sigma (\DB_j^{-1}\AB_j\HB_j \WB_\ast) \vb_\ast +\epsilon_j.
\end{align}
Here, $\{\epsilon_j\}_{j=1}^n$ are independent sub-Gaussian white noises with $\psi_2$ norm $\nu$. Through out this paper, we assume that $\|\WB_\ast\|_2 =  1$. For simplicity, we assume that the learning process for $\WB$ and $\vb$ is disjoint from that for $\ab_j$, or $\ab_{j}$ is fixed and does not need training.

\subsection{Assumptions}
In this paper, we make the following assumptions, which can be easily satisfied with commonly used activation functions in practice, e.g., ReLU, Leaky ReLU, Sigmod and Softplus.
\begin{assumption}\label{assump:1}
	$\sigma$ is a non-trivial increasing function, and is 1-Lipschitz continuous, i.e., $\left| \sigma(x_1)-\sigma(x_2) \right| \le \left|x_1-x_2 \right|, \forall x_1,x_2 \in \RB$.
\end{assumption}
\begin{assumption}\label{assump:2}
	$\norm{\sigma(x)} \le L_\sigma \norm{x}, \forall x \in \RB^n$.
\end{assumption}

\section{Algorithm and Main Results}
\label{sec:analysis}

We study the sample complexity and training dynamics of learning GNNs with one convolution layer. We construct novel algorithms for training GNNs, and further prove that, with high probability, the proposed algorithms with proper initialization and training step size grant a linear convergence to the ground-truth parameters up to statistical precision. Our results apply to general non-trivial, monotonic and
Lipschitz continuous activation functions including ReLU, Leaky ReLU, Sigmod, Softplus and Swish.

\subsection{Algorithm Design}
We present the algorithm in Algorithm~\ref{algo} for NGNN and GGNN with two auxiliary matrices defined as
\begin{align}\label{mat_1}
\Xi_N =  \E\left[ \HB^\top \AB^\top (\DB^{-1})^\top  \sigma (\DB^{-1}\AB\HB ) \right],
\end{align}
and
\begin{align}\label{mat_2}
\Xi_G = \frac{1}{n} \sum_{j=1}^{n} \E\left[ \HB_{j}^\top \AB_{j}^\top (\DB_{j}^{-1})^\top \ab_j^\top \ab_j \sigma (\DB_j^{-1}\AB_j\HB_j ) \right].
\end{align}
It is easy to see that the above two matrices are invertible under Assumption~\ref{assump:1}.
For learning NGNN (similar for GGNN), if we replace $ \Xi_N^{-1} $ by the gradient $\sigma^\prime \left(\DB^{-1}\AB\HB \right) $, we have exact gradients for $\WB$ and $\vb$ with a square loss $\frac{1}{n}\norm{\yb-\sigma (\DB^{-1}\AB\HB \WB) \vb}^2$. However, calculating $\sigma^\prime \left(\DB^{-1}\AB\HB \right) $ for some activation functions could be computationally expensive.  Note that to compute these matrices of \eqref{mat_1} and \eqref{mat_2}, one only needs to make an epoch pass on the dataset. Compared with a regular training by DNNs with tons of epoch passes on the entire dataset to acquire exact gradients, obtaining the two matrices adds very limited burden to the training computation. Hence, the gradient of $\WB$ is inexact in the algorithm and a similar idea is also adopted in analysis of  CNNs~\cite{cao2019tight}. 
\begin{algorithm}[ht]
	\SetAlgoLined
	\KwIn{Training data $\{\HB,\yb\}$, number of iterations $T$, step size $\alpha$, initialization $\WB_0, \vb_0, t=1$.}
	\While{$t < T$}{
		$\bar c  = \begin{cases}
		\DB^{-1}\AB\HB, & \text{NGNN}, \\
	\DB_j^{-1}\AB_j\HB_j, & \text{GGNN}.
		\end{cases} $
		
		$\bar y  = \begin{cases}
		\sigma (c^\prime \WB_t) \vb_t, & \text{NGNN}, \\
		\ab_j \sigma (c^\prime \WB_t) \vb_t, & \text{GGNN}.
		\end{cases} $
		
		$\GB^W_t  = \begin{cases}
		-\frac{1}{n} \Xi_N^{-1} \bar c^\top   \left(\yb -  \bar y \right)\vb_t^\top, & \text{NGNN}, \\
		-\frac{1}{n} \Xi_G^{-1} \sum_{j=1}^{n}  \bar c^\top \ab_j^\top  \left(y_j -\bar y  \right)\vb_t^\top, & \text{GGNN}.
		\end{cases} $
		
		$\gb^v_t = \begin{cases}
		-\frac{1}{n} \sigma(\WB_t^\top\bar c^\top) \left(\yb- \bar{y}\right), & \text{NGNN},\\
		-\frac{1}{n}\sum_{j=1}^n \sigma(\WB_t^\top\bar{c}^\top) \ab_j^\top\left(y_j - \bar{y}\right), &\text{GGNN}.
		\end{cases} $
		
		$\UB_{t+1} = \WB_t - \alpha \GB^W_t$
		
		$\WB_{t+1} = \UB_{t+1}/\|\UB_{t+1}\|_2$
		
		$\vb_{t+1} = \vb_t -\alpha \gb^v_t$
	}
	\KwResult{$\WB_{T}, \vb_T$}
	\caption{Approximate Gradient Descent for Learning GNNs}
	\label{algo}
\end{algorithm}

\subsection{Convergence Analysis}
For the ease of presentation, we first introduce some quantities that we use in this paper.
\begin{align}
&\gamma_1 =\begin{cases}
\sqrt{ \frac{\alpha L_\sigma^2 d_{out}}{2} - 3\alpha^2(d+4d\bar{d})^2L_\sigma^4}, & \text{NGNN},\\
\sqrt{ \frac{\alpha L_\sigma^2 d_{out}}{2} - 12\alpha^2 d^2L_\sigma^4n_{\max}^4}, &  \text{GGNN};
\end{cases}\\
&\gamma_2 =\begin{cases}
\sqrt{\frac{2\alpha (d+4d\bar{d})^2}{d_{out}}+6\alpha^2 (d+4d\bar{d})^2L_\sigma^2}, & \text{NGNN},\\
\sqrt{\frac{8\alpha d^2 n_{\max}^4}{d_{out}}+24\alpha^2 d^2L_\sigma^4n_{\max}^2}, &  \text{GGNN};
\end{cases} \\
&\gamma_3 = 
\sqrt{\frac{2\alpha + 3\alpha ^2 L_\sigma^2 d_{out}}{\gamma^2_1 L_\sigma^2 d_{out}}}, 
\end{align}
where $\bar{d} = \frac{1}{n}\sum_i d_i$ is the average node degree in the graph for NGNN, and $n_{\max} = \max_j n_j$ is the maximum number of nodes in a single graph for GGNN.
Note that the above quantities only depend on the sample and feature properties, model architecture, and learning rate, and are independent of the number of samples $n$. 

Next, we define
\begin{align}D=\max \left\lbrace \norm{\vb_0- \vb_\ast}, \sqrt{\frac{2\gamma_2\norm{\vb_\ast}^2}{\gamma_1}+{\gamma_3}}\right\rbrace,
\end{align}
and
\begin{align} 
\rho = \begin{cases}
\min\left\lbrace \vb_0^\top \vb_\ast
, \frac{L_\sigma^2d_{out}\norm{\vb_\ast}^2}{(d+4d\bar{d})L_\sigma^2+1} \right\rbrace , & \text{NGNN}, \\
\min\left\lbrace \vb_0^\top \vb_\ast
, \frac{L_\sigma^2d_{out}\norm{\vb_\ast}^2}{2dL_\sigma^2n_{\max}^2+1} \right\rbrace , & \text{GGNN}.
\end{cases}
\end{align}
and let $D_0 = D + \norm{\vb_\ast}$. Similarly, the above quantities do not depend on the number of samples $n$.
\subsection{Useful Lemmas}
We provide in this section some useful lemmas to drive our main result. This could also serve as sketch of the proof of the main results and to help improve the presentation. First, two lemmas of the general results for both NGNN and GGNN are presented.
\begin{lemma}\label{graph_level_W_init}
	As the assumptions in Theorem~\ref{convergence_thm} hold, there exists $\eta_W$ such that we have
	\begin{align}
&	\|\WB_{t+1} - \WB_\ast\|_2 - \frac{4\eta_W\left(1+\alpha\rho+\sqrt{1+\alpha\rho}\right)}{\rho}\\
&	\le \frac{1}{\sqrt{1+\alpha\rho}}\left[	\|\WB_{t} - \WB_\ast\|_2-\frac{4\eta_W\left(1+\alpha\rho+\sqrt{1+\alpha\rho}\right)}{\rho}\right].
	\end{align}
\end{lemma}

\begin{lemma}\label{graph_level_v_init}
	As the assumptions in Theorem~\ref{convergence_thm} hold, there exist $\sigma_m$, $\sigma_M$, $L$, and $\eta_v$, such that we have
	\begin{align}
	&\norm{\vb_{t+1}-\vb_\ast}^2 \le \left(1-\frac{\alpha\sigma_m}{2}+3\sigma^2_M\alpha^2\right)\norm{\vb_t-\vb_\ast}^2\\ &+\left(\frac{\alpha L^2}{\sigma_m}+3L^2\alpha^2\right)\norm{\WB_t-\WB_\ast}^2\norm{\vb_\ast}^2+\left(\frac{2\alpha}{\sigma_m}+3\alpha^2\right)\eta_v^2.
	\end{align}
\end{lemma}

Next, we provide results specifically for NGNN and GGNN, respectively. We start by defining the following
\begin{align}
\bar{\GB}^W(\WB,\vb) &= \E_{\HB}\left[{\GB}^W(\WB,\vb) \right] \\
\bar{\gb}^v(\WB,\vb) &= \E_{\HB}\left[ \gb^v(\WB,\vb)  \right]
\end{align}
\subsubsection{Analysis for NGNN}
\begin{lemma}\label{node_level_lips}
	If $n\ge \frac{8\ln \frac{2}{\delta}}{d d_{\min}}$,  with probability at least $1-\delta$
	\begin{align}
	&\frac{\norm{\GB^W(\WB,\vb) -\GB^W(\WB^\prime,\vb) }}{\norm{\WB-\WB^\prime}} \le  D_0^2  \norm{\Xi^{-1}} (d+4d\bar{d}) ,\\
	&\frac{\norm{\GB^W(\WB,\vb) -\GB^W(\WB,\vb^\prime) }}{\norm{\vb-\vb^\prime}} \le 4D_0\norm{\Xi^{-1}}(d+4d\bar{d})L_\sigma,\\
	&\frac{\norm{\gb^v(\WB,\vb)-\gb^v(\WB^\prime,\vb)}}{\norm{\WB-\WB^\prime}} \le 5(d+4d\bar{d})D_0L_\sigma\nn \\
	&\ \ \ \ \ \ \ \ \ \ + \frac{d+4d\bar{d}}{2}+ \nu^2,\\
	&\frac{\norm{\gb^v(\WB,\vb) -\gb^v(\WB,\vb^\prime) }}{\norm{\vb-\vb^\prime}} \le  {L_\sigma^2}(d+4d\bar{d}).
	\end{align}
\end{lemma}
\begin{lemma}\label{node_level_eta}
	If $\sqrt{n\log n}\ge \frac{3}{2}\log \frac{2}{\delta}$, with probability at least $1-\delta$ 
	\begin{align}
	&\norm{\GB^W-\bar\GB^W } \\
	&\le \frac{4}{c}\sqrt{{\frac{\log n}{n}}} \norm{\Xi^{-1}}D_0 \left(D_0\frac{d}{d_{\min}}(1+|\sigma(0)|)+\sqrt{\frac{d}{d_{\min}}}\nu\right) ,\\
	&\norm{\gb^v-\bar\gb^v }\\
	& \le \frac{2}{c}\sqrt{{\frac{\log n}{n}}} \left(D_0\frac{d}{d_{\min}}(1+|\sigma(0)|)^2+\sqrt{\frac{d}{d_{\min}}}(1+|\sigma(0)|)\nu\right) ,
	\end{align}
	for some absolute constant $c$.
\end{lemma}
\subsubsection{Analysis for GGNN}
\begin{lemma}\label{graph_level_lips}
	If $\sum_{j=1}^{n}n_j \ge\frac{8\ln\frac{1}{\delta}}{d} $,  with probability at least $1-\delta$
	\begin{align}
	&\frac{\norm{\GB^W(\WB,\vb) -\GB^W(\WB^\prime,\vb) }}{\norm{\WB-\WB^\prime}} \le 2dD_0^2  \norm{\Xi^{-1}} n^2_{\max}, \\
	&\frac{\norm{\GB^W(\WB,\vb) -\GB^W(\WB,\vb^\prime) }}{\norm{\vb-\vb^\prime}} \le 8D_0d\Xi^{-1}n^2_{\max}L_\sigma,\\
	&\frac{\norm{\gb^v(\WB,\vb)-\gb^v(\WB^\prime,\vb)}}{\norm{\WB-\WB^\prime}} \le 10dD_0L_\sigma n_{\max}^2\nn\\
	&+{d\sqrt{n^3_{\max}}}+{\sqrt{n_{\max}}}\nu^2,\\
	&\frac{\norm{\gb^v(\WB,\vb) -\gb^v(\WB,\vb^\prime) }}{\norm{\vb-\vb^\prime}} \le 2d{L_\sigma^2}n_{\max}^2.
	\end{align}
\end{lemma}
\begin{lemma}\label{graph_level_eta}
	If $\sqrt{n\log n}\ge \frac{3}{2}\log \frac{2}{\delta}$, with probability at least $1-\delta$ 
	\begin{align}
	&\norm{\GB^W-\bar\GB^W }\\
	& \le \frac{4}{c}\sqrt{\frac{\log n}{n}} \norm{\Xi^{-1}} \sqrt{n_{\max}} \left(\sqrt{n_{\max}}2(1+|\sigma(0)|)D_0+\nu\right),\\
	&\norm{\gb^v-\bar\gb^v } \\
	&\le \frac{2}{c}\sqrt{\frac{\log n}{n}} 	\sqrt{n_{\max}}2(1+|\sigma(0)|) \left(	\sqrt{n_{\max}}2(1+|\sigma(0)|)D_0+\nu\right),
	\end{align}
	for some absolute constant $c$.
\end{lemma}

Next, we provide the main results for NGNN and GGNN with the proposed algorithms.
\begin{theorem}\label{convergence_thm}
	Suppose that the initialization $(\WB_0,\vb_0)$ satisfies
	\begin{align}\label{initialization}
	\Tr\left( \WB_{\ast}^\top \WB_0\right) \ge 0, \vb_\ast^\top\vb_0 \ge 0, 
	\end{align}
	and the learning rate $\alpha$ is chosen such that
	\begin{align}
	\alpha \le \begin{cases}
\frac{1}{2(\norm{\vb_0-\vb_\ast}^2+\norm{\vb_\ast}^2)} \wedge \frac{d_{out}}{6(d+4d\bar{d})^2L_\sigma^2}, & \text{NGNN},\\
	\frac{1}{2(\norm{\vb_0-\vb_\ast}^2+\norm{\vb_\ast}^2)} \wedge \frac{d_{out}}{24d^2L_\sigma^2n_{\max}^4}, & \text{GGNN}.
	\end{cases}
	\end{align}
	If the number of samples $n$ is large enough such that for NGNN
	\begin{align}\label{sample_node}
	&\frac{2}{c}\sqrt{\frac{\log n}{n}} \le \frac{\frac{\rho}{16(1+\alpha \rho )\norm{\Xi^{-1}}D_0}\Tr(\WB_\ast^\top\WB_0)\wedge \frac{\rho}{\norm{\vb_\ast}(1+|\sigma(0)|)}}{D_0\frac{d}{d_{\min}^2}(1+|\sigma(0)|)+\frac{d}{d_{\min}}\nu},
	\end{align}
	and for GGNN
	\begin{align}\label{sample_graph}
	&\frac{2}{c}\sqrt{\frac{\log n}{n}} \le \frac{\frac{\rho}{16(1+\alpha \rho )\norm{\Xi^{-1}}D_0}\Tr(\WB_\ast^\top\WB_0)\wedge \frac{\rho}{\norm{\vb_\ast}(1+|\sigma(0)|)}}{ {n_{\max}}(1+|\sigma(0)|)D_0+\sqrt{n_{\max}}\nu},	
	\end{align}
	for some large enough absolute constant $c$,  then with probability at least $1-\delta$ and for large $n$ such that $\sqrt{n\log n}\ge \frac{3}{2} \log \frac{2}{\delta}$ and $n\ge \frac{8\ln \frac{2}{\delta}}{d d_{\min}}$ for NGNN or $\sum_{j=1}^{n}n_j \ge\frac{8\ln\frac{1}{\delta}}{d} $ for GGNN,  we have that
	\begin{align}
	\norm{\WB_{t} - \WB_\ast} \le &\left(\frac{1}{\sqrt{1+\alpha\rho}}\right)^t\norm{\WB_0-\WB_\ast}\\
	&+\eta_W(6\alpha +\frac{8}{\rho})
	\end{align}
	with
	\begin{align}
	&\bar a_W = \frac{4}{c}\sqrt{{\frac{\log n}{n}}}\norm{\Xi^{-1}} D_0,
	\end{align}
	and
	\begin{align}
		&\eta_W = \begin{cases}
		 \bar a_W \left(D_0\frac{d}{d_{\min}}(1+|\sigma(0)|)+\sqrt{\frac{d}{d_{\min}}}\nu\right), & \text{NGNN},\\
	\bar a_W \sqrt{n_{\max}} \left(\sqrt{n_{\max}}(1+|\sigma(0)|)D_0+\nu\right), & \text{GGNN};
		\end{cases}
	\end{align}
	and
	\begin{align}\
&	\norm{\vb_{t}-\vb_\ast} \le \left(\sqrt{1-\gamma^2_1}\right)^t\norm{\vb_0-\vb_\ast}\nn\\ &+\gamma_2\norm{\WB_0-\WB_{\ast}}\norm{\vb_\ast} \sqrt{t}\left(\sqrt{1-\gamma^2_1}\vee \sqrt{\frac{1}{{1+\alpha\rho}}}\right)^{t-1}\nn\\
	&+\gamma_3\eta_v+\frac{\gamma_2}{\gamma_1}(6\alpha +\frac{8}{\rho})\norm{\vb_\ast}\eta_W
	\end{align}
	with 
	\begin{align}
	&\bar{a}_v = 	\frac{2}{c}\sqrt{{\frac{\log n}{n}}},
	\end{align}
	and
	\begin{align} 
		&\eta_v = \nn\\
		&\begin{cases}
	\bar{a}_v \left(D_0\frac{d}{d_{\min}}(1+|\sigma(0)|)^2+\sqrt{\frac{d}{d_{\min}}}(1+|\sigma(0)|)\nu\right), & \text{NGNN},\\
	\bar{a}_v	\sqrt{n_{\max}}(1+|\sigma(0)|) \left(	\sqrt{n_{\max}}(1+|\sigma(0)|)D_0+\nu\right), & \text{GGNN}.
		\end{cases}  
	\end{align}
\end{theorem}
\begin{remark}
Essentially, it is shown in Theorem~\ref{convergence_thm} that the proposed algorithm can provably learn GNNs with a linear convergence to the true parameters of GNNs up to a statistical error. The error is governed by 
the number of training samples $n$, and approaches 0 when $n$ is infinitely large. Therefore, exact convergence of the algorithm  is guaranteed for large $n$, and this result is intuitively desirable and expected. 
\end{remark}
\begin{remark}
Theorem~\ref{convergence_thm} requires that the initialization satisfies \eqref{initialization}, which may not necessarily be satisfied by a random initialization. One solution is to try out all 4 sign combinations of 
\[(\WB_0,\vb_0) \in \{(\WB,\vb), (-\WB,\vb),(\WB,-\vb),(-\WB,-\vb)\} \] as suggested in \cite{du2017gradient}. This 
guarantees that the initialization condition of~\eqref{initialization} can be satisfied, and  further the convergence of the algorithm.
\end{remark}
\begin{remark}
The conditions on the number of sampels in~\eqref{sample_node} and~\eqref{sample_graph} pose sufficient conditions on the number of samples $n$ for learning  NGNN and GGNN, respectively. Intuitively, an efficient learning algorithm needs sufficiently many learning samples, and we provide such analytical results in this regard by the two conditions. 
\end{remark}

It would also be interesting to explore the impact of the structures of the graphs on the convergence, e.g., the convergence could be possibly optimized with respect to the spectral properties fo the graphs . Although we include node degrees, number of nodes in the derived convergence, which also relate to the structures of the graphs, we leave such a work to the interested community and our future work due to the length limit of this paper.

While the general differences between the GNNs and CNNs are discussed in Related Work, we emphasize hereby that the resulting difference for technical analysis is even more significant. This is due to unordered and overlapping neighbors in GNNs. There has been limited work on CNNs with overlapping structures (see \cite{cao2019tight} and the references therein). For technical analysis, non-overlapping CNNs yields independent data variables for each convolution patch and then tractable concentration bounds like Lemma 5.2 in \cite{cao2019tight}, which leads to easier and simpler analysis compared with the analysis in this paper.  On the contrary, Also, the analysis over graph for GNNs adds much more difficulties compared with that of CNNs. We develop new techniques in the theoretical analysis to tackle challenges from dependent variables (from overlapping neighbors) in graphs (it is equivalently grid for CNNs), and to design provable efficient algorithm on top of it. Please refer to  Lemmas 3-10 in the supplementary material for related detail.  Thus, the analyses for CNNs are special cases in our paper. 

\section{Training Dynamics}
In the previous section, we show that the outputs of the proposed algorithms are provably convergent to underlying true parameters. At this point, we have not fully understood the training dynamics of the estimated parameters $ \WB_{t}$ and $ \vb_{t}$. A missing building brick to the analysis would be the proposition of a compact subspace that the training sequences of $ \WB_{t}$ and $ \vb_{t}$ lie within.

Toward this goal, we first define the space
\begin{align}
&\WC=\left\lbrace \WB: \norm{\WB}=1, \Tr(\WB_\ast^\top\WB) \ge {\Tr\left(\WB^\top_\ast \WB_0\right)}/2 \right\rbrace,
\end{align}
and
\begin{align} 
&\VC = \left\lbrace \vb: \norm{\vb-\vb_\ast} \le D, \vb^\top_\ast\vb \ge \rho \right\rbrace .
\end{align}
We then have the following theorem on the stability of the training procedure.
\begin{theorem}\label{training_dynamics}
	Suppose the assumptions in Theorem~\ref{convergence_thm} hold, then the training dynamics are confined to
	\begin{align}
	\left\lbrace \WB_{t}, \vb_{t}\right\rbrace \in \WC \times \VC.
	\end{align}
\end{theorem}

Theorem~\ref{training_dynamics} states that the training process of the proposed learning algorithm is provably stable, in addition to Theorem~\ref{convergence_thm} which grants convergence of the training outputs.

\section{Experimental Results}
We provide numerical experiments to support and validate our theoretical analysis. We test Algorithm~\ref{algo} for NGNN and  GGNN tasks, respectively. Different activation functions of ReLU, Leaky ReLU, Sigmoid, Tanh and Swish are used in the networks, and the results with respect to the distance between the estimated and the true parameters, i.e., $\norm{\WB_\ast-\WB_t}$ and $\norm{\vb_\ast-\vb_t}$, versus the number of training epochs are reported. Specifically, for the networks with Leaky ReLU, we show results for two different slope parameter of 0.2 and 0.05. We choose $d=2$ and $d_{out} = 1$, and set the variance $\nu$ to 0.04. We generate $\WB_{\ast}$ from unit sphere with a normalized Gaussian matrix, and generate $\vb_\ast$ as a standard Gaussian vector. 
The nodes in the graphs are probabilistically connected according to the distribution of Bernoulli(0.5).
\subsection{NGNN Tasks}
For node-level tasks, we have one graph and 1000 nodes. 
\begin{figure}[!h]
	\centering
\caption{Training dynamics for NGNN with different activation functions.}
	\subfloat[ReLU]{\label{nfigur:1}\includegraphics[width=0.5\linewidth]{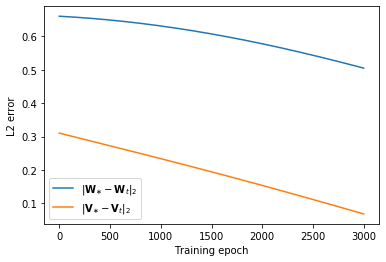}}
	\subfloat[Leaky ReLU (0.2)]{\label{nfigur:2}\includegraphics[width=0.5\linewidth]{node_relu_521dot04.png}}
	\\
	\subfloat[Leaky ReLU (0.05)]{\label{nfigur:3}\includegraphics[width=0.5\linewidth]{node_relu_521dot04.png}}
	\subfloat[Sigmoid]{\label{nfigur:4}\includegraphics[width=0.5\linewidth]{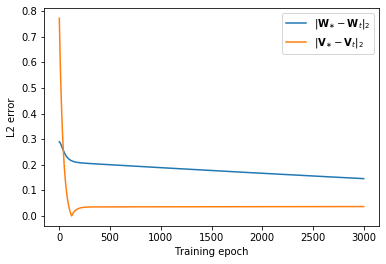}}\\
	\subfloat[Tanh]{\label{nfigur:5}\includegraphics[width=0.5\linewidth]{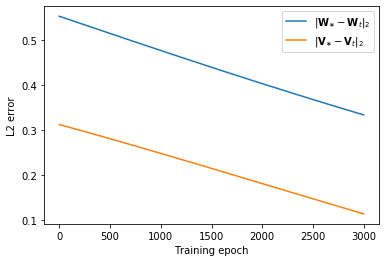}}
	\subfloat[Swish]{\label{nfigur:6}\includegraphics[width=0.5\linewidth]{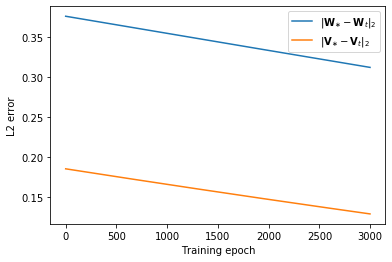}}	\label{fig-node}
\end{figure}

 The learning rate $\alpha$ is chosen as 0.1. As presented in Figure~\ref{fig-node}, we can see clearly the stable training of the GNN and the efficiency of the algorithm. It is show that $\WB$ converges slower than $\vb$. Another interesting finding is that when Sigmoid is used as the activation function, the training of $\vb$ very quickly converges to the optimum, and then lingers at the neighborhood of $\vb_\ast$. This observation validates our theoretical result that the convergence is up to a statistical error. Also due to this reason, the actual convergence might be slightly slower than linear rate, which however is inevitable because of the statistical model that we consider. 

\subsection{GGNN Tasks}
\begin{figure}[!h]
	\centering
\caption{Training dynamics for GGNN with different activation functions.}
	\subfloat[ReLU]{\label{gfigur:1}\includegraphics[width=0.5\linewidth]{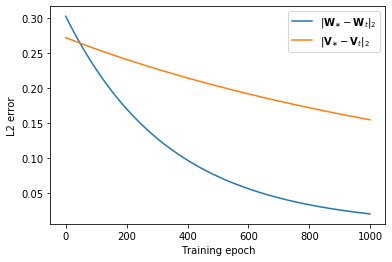}}
	\subfloat[Leaky ReLU (0.2)]{\label{gfigur:2}\includegraphics[width=0.5\linewidth]{graph_relu_521dot04.png}}
	\\
	\subfloat[Leaky ReLU (0.05)]{\label{gfigur:3}\includegraphics[width=0.5\linewidth]{graph_relu_521dot04.png}}
	\subfloat[Sigmoid]{\label{gfigur:4}\includegraphics[width=0.5\linewidth]{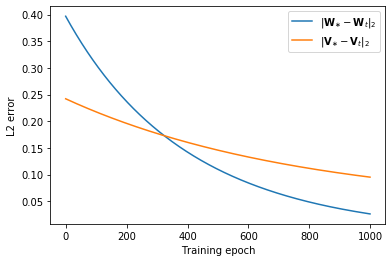}}\\
	\subfloat[Tanh]{\label{gfigur:5}\includegraphics[width=0.5\linewidth]{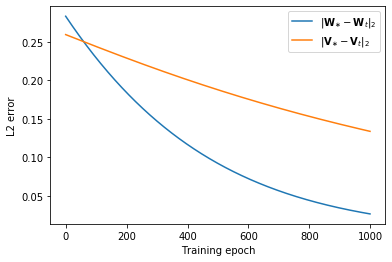}}
	\subfloat[Swish]{\label{gfigur:6}\includegraphics[width=0.5\linewidth]{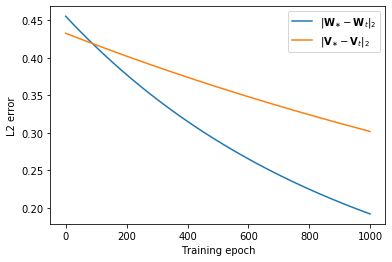}}	\label{fig-graph}
\end{figure}
For GGNN tasks, we have 1000 graphs and each graph has 5 nodes. The learning rate $\alpha$ is chosen as 0.005. As presented in Figure~\ref{fig-graph}, we can see clearly the training dynamics of the GNN and the efficiency of the algorithm. The curves are smooth, indicating that the training is very stable. By looking at the slope of the curves, for NGNN, we find that $\WB$ converges faster than $\vb$, which is very different from the NGNN task.

\section{Conclusion}
In this paper, we developed the first provably efficient algorithm for learning GNNs with one hidden layer for node information convolution. We investigated two types of GNNs, namely, node-level and graph-level GNNs. Our results provided a comprehensive framework and a set of tools for designing and analyzing GNNs. More importantly, our results only rely on mild condition on the activation functions, which can be satisfied by a wide range of activation functions used in practice. We constructed our training algorithm using the idea of approximate gradient descent, and proved that it recovers the true parameters of the teacher network with a linear convergence rate. How fast the algorithm converges depends on various parameters of the algorithm, which was also analytically characterized. 

\newpage
\section*{Broader Impact}
GNN's success has been demonstrated empirically in a wide range of applications, including social networks, recommender systems, knowledge graphs, protein interface networks, and generative models. Our work is the first to theoretically understand why GNNs work well in practice, and our algorithm converges to underlying true parameters of the teacher network linearly. Our results and tools developed in this paper provide a general framework to develop further theoretical understandings of GNNs. 

There are many benefits to using the proposed algorithm, such as guaranteeing the convergence and learnability in decision-critical applications. This can help mitigate fairness, privacy and safety risks. The potential risks of increasing convergence and learnability include: (i) the risk of an undue trust in models, (ii) if the guarantee of learnability means the algorithm may now be used by those with lower levels of domain or ML expertise, this could increase the risk of the model or its outputs being used incorrectly.

\bibliographystyle{aaai21}
\bibliography{lqw}
\newpage
\onecolumn
\newpage
\appendix
\section{Proofs of the Main Results}
\subsection{General Results for NGNN and GGNN}
\begin{lemma}\label{graph_level_W_init}
	As the assumptions in Theorem~\ref{convergence_thm} hold, there exists $\eta_W$ such that we have
	\begin{align}
	\|\WB_{t+1} - \WB_\ast\|_2 - \frac{4\eta_W\left(1+\alpha\rho+\sqrt{1+\alpha\rho}\right)}{\rho}\le \frac{1}{\sqrt{1+\alpha\rho}}\left[	\|\WB_{t} - \WB_\ast\|_2-\frac{4\eta_W\left(1+\alpha\rho+\sqrt{1+\alpha\rho}\right)}{\rho}\right].
	\end{align}
\end{lemma}

\begin{lemma}\label{graph_level_v_init}
	As the assumptions in Theorem~\ref{convergence_thm} hold, there exist $\sigma_m$, $\sigma_M$, $L$, and $\eta_v$, such that we have
	\begin{align}
	\norm{\vb_{t+1}-\vb_\ast}^2 \le \left(1-\frac{\alpha\sigma_m}{2}+3\sigma^2_M\alpha^2\right)\norm{\vb_t-\vb_\ast}^2 +\left(\frac{\alpha L^2}{\sigma_m}+3L^2\alpha^2\right)\norm{\WB_t-\WB_\ast}^2\norm{\vb_\ast}^2+\left(\frac{2\alpha}{\sigma_m}+3\alpha^2\right)\eta_v^2.
	\end{align}
\end{lemma}

\subsection{Analysis for NGNN}
We first give a few definitions as follows
\begin{align}
\bar{\GB}^W(\WB,\vb) &= \E\left[{\GB}^W(\WB,\vb) \right] =\frac{1}{n}\left(  \WB\vb\vb^\top-\WB^\ast\vb^\ast\vb^\top\right), \\
\phi(\WB,\WB^\prime) &= \E_{\HB\sim \NC(0,1)}\left[\frac{1}{n} \left( \sigma (\DB^{-1}\AB\HB\WB)\right) ^\top\sigma (\DB^{-1}\AB\HB\WB^\prime)\right],\\
\bar{\gb}^v(\WB,\vb) &= \E\left[ \gb^v(\WB,\vb)  \right]=\left( \phi(\WB,\WB)\vb-\phi(\WB,\WB_\ast)\vb_\ast\right) \\
&=\left( \phi(\WB,\WB)(\vb-\vb_\ast)+\left(\phi(\WB,\WB)-\phi(\WB,\WB_\ast))\right)\vb_\ast\right) ,
\\
\phi_{t,t} &= \phi(\WB_t,\WB_t), \phi_{t,\ast} = \phi(\WB_t,\WB_\ast),
\end{align}
\begin{align}
\bar{\vb}_{t+1} = \vb_t - \alpha\bar{\gb}^v_t.
\end{align}
We replace the parentheses of $(\WB,\vb)$ in above definitions by subscript $t$ in proper places to denote $(\WB_t,\vb_t)$ for the ease of presentation.
\begin{lemma}\label{node_level_lips}
	If $n\ge \frac{8\ln \frac{2}{\delta}}{d d_{\min}}$,  with probability at least $1-\delta$
	\begin{align}
	&\frac{\norm{\GB^W(\WB,\vb) -\GB^W(\WB^\prime,\vb) }}{\norm{\WB-\WB^\prime}} \le  D_0^2  \norm{\Xi^{-1}} (d+4d\bar{d}) ,\\
	&\frac{\norm{\GB^W(\WB,\vb) -\GB^W(\WB,\vb^\prime) }}{\norm{\vb-\vb^\prime}} \le 4D_0\norm{\Xi^{-1}}(d+4d\bar{d})L_\sigma,\\
	&\frac{\norm{\gb^v(\WB,\vb)-\gb^v(\WB^\prime,\vb)}}{\norm{\WB-\WB^\prime}} \le 5(d+4d\bar{d})D_0L_\sigma + \frac{d+4d\bar{d}}{2}+ \nu^2,\\
	&\frac{\norm{\gb^v(\WB,\vb) -\gb^v(\WB,\vb^\prime) }}{\norm{\vb-\vb^\prime}} \le  {L_\sigma^2}(d+4d\bar{d}).
	\end{align}
\end{lemma}
\begin{lemma}\label{node_level_eta}
	If $\sqrt{n\log n}\ge \frac{3}{2}\log \frac{2}{\delta}$, with probability at least $1-\delta$ 
	\begin{align}
	&\norm{\GB^W-\bar\GB^W } \le \frac{4}{c}\sqrt{{\frac{\log n}{n}}} \norm{\Xi^{-1}}D_0 \left(D_0\frac{d}{d_{\min}}(1+|\sigma(0)|)+\sqrt{\frac{d}{d_{\min}}}\nu\right) ,\\
	&\norm{\gb^v-\bar\gb^v } \le \frac{2}{c}\sqrt{{\frac{\log n}{n}}} \left(D_0\frac{d}{d_{\min}}(1+|\sigma(0)|)^2+\sqrt{\frac{d}{d_{\min}}}(1+|\sigma(0)|)\nu\right) ,
	\end{align}
	for some absolute constant $c$.
\end{lemma}

\subsection{Analysis for GGNN}
We first give a few definitions as follows
\begin{align}
\bar{\GB}^W(\WB,\vb) &=\E\left[ {\GB}^W(\WB,\vb)\right] = \WB\vb\vb^\top-\WB^\ast\vb^\ast\vb^\top,\\
\phi_j(\WB,\WB^\prime) &= \E_{\HB_{j}\sim \NC(0,1)}\left[\left(\ab_j \sigma (\DB_j^{-1}\AB_j\HB_j\WB)\right)^\top\ab_j \sigma (\DB_j^{-1}\AB_j\HB_j\WB^\prime)\right],\\
\phi(\WB,\WB^\prime) &= \frac{1}{n}\sum_{j=1}^n 	\phi_j(\WB,\WB^\prime),\\
\bar{\gb}^v(\WB,\vb) &= \E\left[ \gb^v(\WB,\vb)  \right]=	\frac{1}{n}\sum_{j=1}^n\phi_j(\WB,\WB)\vb-\frac{1}{n}\sum_{j=1}^n\phi_j(\WB,\WB_\ast)\vb_\ast\\
&=	\phi(\WB,\WB)\vb-\phi(\WB,\WB_\ast)\vb_\ast\\
&=\phi(\WB,\WB)(\vb-\vb_\ast)+\left(\phi(\WB,\WB)-\phi(\WB,\WB_\ast)\right)\vb_\ast,\\
\phi_{t,t} &= \phi(\WB_t,\WB_t), \phi_{t,\ast} = \phi(\WB_t,\WB_\ast).
\end{align}
We replace the parentheses of $(\WB,\vb)$ by subscript $t$ in proper places to denote $(\WB_t,\vb_t)$ for the ease of presentation.

\begin{lemma}\label{graph_level_lips}
	If $\sum_{j=1}^{n}n_j \ge\frac{8\ln\frac{1}{\delta}}{d} $,  with probability at least $1-\delta$
	\begin{align}
	&\frac{\norm{\GB^W(\WB,\vb) -\GB^W(\WB^\prime,\vb) }}{\norm{\WB-\WB^\prime}} \le 2dD_0^2  \norm{\Xi^{-1}} n^2_{\max}, \\
	&\frac{\norm{\GB^W(\WB,\vb) -\GB^W(\WB,\vb^\prime) }}{\norm{\vb-\vb^\prime}} \le 8D_0d\Xi^{-1}n^2_{\max}L_\sigma,\\
	&\frac{\norm{\gb^v(\WB,\vb)-\gb^v(\WB^\prime,\vb)}}{\norm{\WB-\WB^\prime}} \le 10dD_0L_\sigma n_{\max}^2+{d\sqrt{n^3_{\max}}}+{\sqrt{n_{\max}}}\nu^2,\\
	&\frac{\norm{\gb^v(\WB,\vb) -\gb^v(\WB,\vb^\prime) }}{\norm{\vb-\vb^\prime}} \le 2d{L_\sigma^2}n_{\max}^2.
	\end{align}
\end{lemma}
\begin{lemma}\label{graph_level_eta}
	If $\sqrt{n\log n}\ge \frac{3}{2}\log \frac{2}{\delta}$, with probability at least $1-\delta$ 
	\begin{align}
	&\norm{\GB^W-\bar\GB^W } \le \frac{4}{c}\sqrt{\frac{\log n}{n}} \norm{\Xi^{-1}} \sqrt{n_{\max}} \left(\sqrt{n_{\max}}2(1+|\sigma(0)|)D_0+\nu\right),\\
	&\norm{\gb^v-\bar\gb^v } \le \frac{2}{c}\sqrt{\frac{\log n}{n}} 	\sqrt{n_{\max}}2(1+|\sigma(0)|) \left(	\sqrt{n_{\max}}2(1+|\sigma(0)|)D_0+\nu\right),
	\end{align}
	for some absolute constant $c$.
\end{lemma}

\subsection{Useful Lemmas and Results}
\begin{lemma}\label{node_lip}
	If $n\ge \frac{8\ln \frac{1}{\delta}}{d d_{\min}}$ with $d_{\min}$ being the smallest node degree in the graph, then with probability at least $1-\delta$ for NGNN
	\begin{align} 
	\frac{1}{n}\norm{ \DB^{-1}\AB\HB}^2
	\le d+4d\bar{d},
	\end{align}
	where $\bar{d}=\frac{\sum_{i=1}^n\frac{1}{d_i}}{n}$.
\end{lemma}
\begin{lemma}\label{lip}
	If $\sum_{j=1}^{n}n_j \ge\frac{8\ln\frac{1}{\delta}}{d} $, with probability at least $1-\delta$ for GGNN
	\begin{align}
	\frac{1}{n}  \sum_{j=1}^{n} \norm{\HB_{j}^\top \AB_{j}^\top (\DB_{j}^{-1})^\top}\norm{ \ab_j}^2\norm{ \DB_j^{-1}\AB_j\HB_j  }\le 2dn^2_{\max}.
	\end{align}
\end{lemma}

We provide several concentration results for random variables, and we do not provide corresponding proofs as the results are commonly acknowledged. 

\begin{lemma}[Hoeffding Inequality for Chi-square Distribution]\label{chi_hoeffding}
	$z_k\sim \NC(0,1)$, for all $t\in[0,1]$ then the probability
	\begin{align}
	P\left(\sum_{k=1}^{n}z_k^2 \le (t+1)n\right)\ge 1-\exp(-\frac{nt^2}{8}).
	\end{align}
\end{lemma}

\begin{lemma}\label{node_SE_concentration}
	Let $X_1,\ldots,X_n$ be independent sub-exponential random variables such that $\EC X_i=\mu_i$ and $X_i \in SE(\nu_i^2, \alpha_i)$. Then\begin{align}
		\sum_{i=1}^n(X_i-\mu_i) \in SE(\sum_{i=1}^n\nu_i^2, \max_i \alpha_i).
	\end{align}
	In particular, denote $\nu_\ast^2 = \sum_{i=1}^n\nu_i^2, \alpha_\ast = \max_i \alpha_i$. Then,
	\begin{align}
		P\left(\frac{1}{n} 	\sum_{i=1}^n(X_i-\mu_i)  \ge t \right)\le \begin{cases}
			\exp\left(-\frac{nt^2}{2\nu_\ast^2 }\right)	, & \text{if}\  0<nt<\frac{\nu_\ast^2 }{\alpha_\ast} \\
			\exp\left(-\frac{nt}{2\alpha_\ast}\right), & \text{otherwise}
		\end{cases}.
	\end{align}
\end{lemma}

\begin{lemma}[\cite{honorio2014tight}]\label{node_SE}
	Let $s$ be a sub-Gaussian variable with parameter $\sigma_s$ and mean $\mu_s = \EB s$. Then, $s^2\in SE(4\sigma_s^2,4\sigma_s^2)$.
\end{lemma}

\begin{lemma}
	A random variable $x$ is called sub-Gaussian, if\begin{align}
	P(|x|\ge t)\le 2e^{-ct^2},
	\end{align}
	for some $c>0$ and $\|x\|_{\psi_2}\le c$. Then let $x_1,\ldots,x_n$ be zero-mean independent sub-Gaussian random variables, the general version of the Hoeffding's inequality states that
	\begin{align}
	P\left(|\sum_{i=1}^{n}x_i|\ge t \right) \le 2\exp\left(-\frac{ct^2}{\sum_{i=1}^{n}\|x_i\|^2_{\psi_2}}\right).
	\end{align}
\end{lemma}

\begin{result}\label{eigen}
	For node-level GNN tasks, 
	based on Lemma \ref{node_lip}, the largest eigenvalue of $\phi_{t,t}$  can be well bounded such that $\sigma_M \le (d+4d\bar{d})L_\sigma^2 $. Similarly, we have $L\le (d+4d\bar{d})L_\sigma$.
	By Theorem 2.2 in \cite{yaskov2014}, almost surely, the smallest eigenvalues of $\phi_{t,t}$ and $\phi_{t,\ast}$ are lower bounded by an absolute constant, which we denote here by $\sigma_m \le L_\sigma^2d_{out}$ and $\sigma^\prime_m \le L_\sigma^2d_{out}$, respectively . 
	
	For graph-level GNN tasks, based on Lemma \ref{lip}, the largest eigenvalue of $\phi_{t,t}$  can be well bounded such that $\sigma_M \le 2dL_\sigma^2 n_{\max}^2$. Similarly, we have {$L \le 2dn^2_{\max}L_\sigma$}.
	By Theorem 2.2 in \cite{yaskov2014}, almost surely, the smallest eigenvalues of $\phi_{t,t}$ and $\phi_{t,\ast}$ are lower bounded by an absolute constant, which we denote here by $\sigma_m \le L_\sigma^2d_{out}$ and $\sigma^\prime_m \le L_\sigma^2d_{out}$, respectively.

	We will use the above equality in the paper, and the results derived still hold.
\end{result}

\subsection{Proofs of Lemmas and Theorems}
In this section, we provide proofs of the above lemmas and the main theorems.
\subsubsection*{Proof of Lemma~\ref{graph_level_W_init}}
\begin{proof}
	Define 
	\begin{align}
	\bar{\UB}_{t+1} = \WB_t - \alpha\bar \GB^W_t = \WB_t(\IB-\alpha \vb_t\vb_t^\top)+\alpha \WB^\ast\vb^\ast\vb_t^\top,
	\end{align}
	and
	\begin{align}
	\hat{\UB} = \bar{\UB}_{t+1}/\|\bar{\UB}_{t+1}\|_2.
	\end{align}
	If {{$\IB-\alpha \vb_t\vb_t^\top \ge 0 $}} and {$\vb_t^\top\vb^\ast \ge \rho$}, $\hat{\UB}$ is closer to $\WB^\ast$ than $\UB^\prime$, i.e., $\|\hat{\UB}-\WB^\ast\|_2 \le \|{\UB^\prime}-\WB^\ast\|_2$ where $\UB^\prime = \frac{\WB_t + \alpha \rho \WB^\ast}{\|\WB_t + \alpha \rho \WB^\ast\|_2}$. Therefore,
	\begin{align}
	1-\frac{1}{2}\Tr\left(\left(\WB_{\ast}-\hat{\UB}\right)^\top \left(\WB_{\ast}-\hat{\UB}\right) \right) \ge 1-\frac{1}{2}\Tr\left(\left(\WB_{\ast}-{\UB}^\prime\right)^\top \left(\WB_{\ast}-{\UB}^\prime\right) \right).
	\end{align}
	Thus, we have
	\begin{align}
	\Tr\left(\frac{1}{2}\WB^\top_\ast\WB_\ast+\frac{1}{2}\hat{\UB}^\top\hat{\UB}-\frac{1}{2}\left(\WB_{\ast}-\hat{\UB}\right)^\top \left(\WB_{\ast}-\hat{\UB}\right) \right) \\
	\ge \Tr\left(\frac{1}{2}\WB^\top_\ast\WB_\ast+\frac{1}{2}{\UB^\prime}^\top{\UB}^\prime-\frac{1}{2}\left(\WB_{\ast}-{\UB}^\prime\right)^\top \left(\WB_{\ast}-{\UB}^\prime\right) \right),
	\end{align}
	which leads to
	\begin{align}
	&\Tr\left(\frac{1}{2}\WB^\top_\ast\WB_\ast+\frac{1}{2}\hat{\UB}^\top\hat{\UB}-\frac{1}{2}\left(\WB_{\ast}-\hat{\UB}\right)^\top \left(\WB_{\ast}-\hat{\UB}\right) \right) \\
	&\ge \Tr\left(\WB^\top_\ast {\UB}^\prime \right) = \Tr\left(\WB^\top_\ast \frac{\WB_t + \alpha \rho \WB^\ast}{\|\WB_t + \alpha \rho \WB^\ast\|_2} \right) \\
	&\ge \Tr\left(\WB^\top_\ast \frac{\WB_t + \alpha \rho \WB^\ast}{1 + \alpha \rho } \right)\\
	&=\Tr\left(\frac{\alpha \rho \WB^\top_\ast \WB^\ast}{1+\alpha \rho }+ \frac{1}{1+\alpha\rho}\left(\frac{1}{2}\WB^\top_\ast\WB_\ast+\frac{1}{2}{\WB_t}^\top\WB_t-\frac{1}{2}\left(\WB_{\ast}-\WB_t\right)^\top \left(\WB_{\ast}-\WB_t\right)\right)\right).
	\end{align}
	Hence, we have
	\begin{align}
	&	\Tr\left(\frac{1}{2}\WB^\top_\ast\WB_\ast+\frac{1}{2}\hat{\UB}^\top\hat{\UB}-\frac{1}{2}\left(\WB_{\ast}-\hat{\UB}\right)^\top \left(\WB_{\ast}-\hat{\UB}\right) \right) \\
	& \ge \Tr\left(\frac{\alpha \rho \WB^\top_\ast \WB^\ast}{1+\alpha \rho }+ \frac{1}{1+\alpha\rho}\left(\frac{1}{2}\WB^\top_\ast\WB_\ast+\frac{1}{2}{\WB_t}^\top\WB_t-\frac{1}{2}\left(\WB_{\ast}-\WB_t\right)^\top \left(\WB_{\ast}-\WB_t\right)\right)\right).
	\end{align}
	Rearranging the terms yields
	\begin{align}
	&\Tr\left( \left(\WB_{\ast}-\hat{\UB}\right)^\top \left(\WB_{\ast}-\hat{\UB}\right) \right)	\\
	&\le \Tr\left(\WB^\top_\ast\WB_\ast +  \hat{\UB}^\top\hat{\UB} -\frac{1+2\alpha\rho}{1+\alpha\rho}\WB^\top_\ast\WB_\ast - \frac{1}{1+\alpha\rho}{\WB_t}^\top\WB_t+\frac{1}{1+\alpha\rho}\left(\WB_{\ast}-\WB_t\right)^\top \left(\WB_{\ast}-\WB_t\right)\right),
	\end{align}
	which is equivalent to
	\begin{align}
	&	\|\WB_{\ast}-\hat{\UB}\|_2^2 \le 1+1-\frac{1+2\alpha\rho}{1+\alpha\rho}-\frac{1}{1+\alpha\rho}+\frac{1}{1+\alpha\rho}\|\WB_{\ast}-\WB_t\|_2^2 = \frac{1}{1+\alpha\rho}\|\WB_{\ast}-\WB_t\|_2^2.
	\end{align}
	Therefore, we obtain
	\begin{align}
	\|\WB_{\ast}-\hat{\UB}\|_2 \le \frac{\|\WB_{\ast}-\WB_t\|_2}{\sqrt{1+\alpha\rho}}.
	\end{align}
	As we have {in Lemmas~\ref{node_level_eta} and~\ref{graph_level_eta} that
		\begin{align}
		\|\UB_{t+1}-\bar{\UB}_{t+1}\|_2 = \alpha \norm{\GB^W_t-\bar{\GB}_t^W} \le \alpha \eta_W,
		\end{align}
	} it results in
	\begin{align}
	\|\WB_{t+1} - \hat{\UB}\|_2 \le \frac{2\alpha \eta_W}{\|\bar{\UB}_{t+1}\|_2}.
	\end{align}
	{
		By Theorem~\ref{training_dynamics}, we have $\Tr(\WB^\top_\ast\WB_t) \ge  \0$. Hence,
		\begin{align}
		\|\bar{\UB}_{t+1}\|_2 = \|\WB_t(\IB-\alpha \vb_t\vb_t^\top)+\alpha \WB^\ast\vb^\ast\vb_t^\top\|_2 \ge 1-\alpha\|\vb_t\|^2_2 \ge 1-\alpha D_0^2 \ge \frac{1}{2}.
		\end{align}
	}
	Thus, it yields
	\begin{align}
	\|\WB_{t+1} - \hat{\UB}\|_2 \le {4\alpha \eta_W}.
	\end{align}
	By triangle inequality, 
	\begin{align}
	\|\WB_{t+1} - \WB_\ast\|_2 \le \|\WB_{\ast} - \hat{\UB}\|_2+\|\WB_{t+1} - \hat{\UB}\|_2 \le {4\alpha \eta_W} + \frac{\|\WB_{\ast}-\WB_t\|_2}{\sqrt{1+\alpha\rho}}, 
	\end{align}
	which results in
	\begin{align}
	\|\WB_{t+1} - \WB_\ast\|_2 - \frac{4\eta_W\left(1+\alpha\rho+\sqrt{1+\alpha\rho}\right)}{\rho}\le \frac{1}{\sqrt{1+\alpha\rho}}\left[	\|\WB_{t} - \WB_\ast\|_2-\frac{4\eta_W\left(1+\alpha\rho+\sqrt{1+\alpha\rho}\right)}{\rho}\right].
	\end{align}
\end{proof}

\subsubsection*{Proof of Lemma~\ref{graph_level_v_init}}
\begin{proof}
First ,we define \begin{align}
	\bar{\vb}_{t+1} = \vb_t - \alpha\bar{\gb}^v_t.
	\end{align}
	Then, it directly follows
	\begin{align}
	\norm{\bar{\vb}_{t+1}-\vb_\ast}=\norm{\left(\IB-\alpha \phi_{t,t}\right)\left(\vb_t-\vb_\ast\right)-\alpha\left(\phi_{t,t}-\phi_{t,\ast}\right)\vb_\ast}.
	\end{align}
	As we have in Lemmas~\ref{node_level_eta} and~\ref{graph_level_eta} that
	\begin{align}
	\norm{\vb_{t+1}-\bar{\vb}_{t+1}} = \alpha \norm{\gb_t^v-\bar{\gb}_t^v} \le \alpha \eta_v.
	\end{align}
	By triangle inequality, we have
	\begin{align}
	\norm{\vb_{t+1}-\vb_\ast} \le \norm{\IB-\alpha \phi_{t,t}}\norm{\vb_t-\vb_\ast} + \alpha\norm{\phi_{t,t}-\phi_{t,\ast}}\norm{\vb_\ast} + \alpha \eta_v
	\end{align}
	According to Lemma~\ref{node_level_eta} and Lemma~\ref{graph_level_eta}, it can be written as
	\begin{align}\label{phi_L}
	\norm{\phi_{t,t}-\phi_{t,\ast}} \le L\norm{\WB-\WB_\ast},
	\end{align}
	where $L$ is given in Result~\ref{eigen}.
	Thus, we can write
	\begin{align}
	\norm{\vb_{t+1}-\vb_\ast} \le \norm{\IB-\alpha \phi_{t,t}}\norm{\vb_t-\vb_\ast} + \alpha L\norm{\WB-\WB_\ast}\norm{\vb_\ast} + \alpha \eta_v.
	\end{align}
	Additionally, we also have
	\begin{align}
	(\vb_t-\vb_\ast)^\top\gb_t^v &\ge (\vb_t-\vb_\ast)^\top\bar\gb_t^v-\eta_v\norm{\vb_t-\vb_\ast}\\
	&=(\vb_t-\vb_\ast)^\top\phi_{t,t}(\vb_t-\vb_\ast)+(\vb_t-\vb_\ast)^\top\left(\phi_{t,t} -\phi_{t,\ast} \right)\vb_\ast-\eta_v\norm{\vb_t-\vb_\ast}\\
	&	\ge \sigma_m\norm{\vb_t-\vb_\ast}^2 - L\norm{\WB_t-\WB_\ast}\norm{\vb_t-\vb_\ast}\norm{\vb_\ast}-\eta_v\norm{\vb_t-\vb_\ast}\\
	& \ge \sigma_m\norm{\vb_t-\vb_\ast}^2-\frac{1}{2}\left(\frac{L^2}{\sigma_m}\norm{\WB_t-\WB_\ast}^2\norm{\vb_\ast}^2+\sigma_m\norm{\vb_t-\vb_\ast}^2\right)-\eta_v\norm{\vb_t-\vb_\ast}\\
	&\ge\frac{\sigma_m}{2} \norm{\vb_t-\vb_\ast}^2-\frac{L^2}{2\sigma_m}\norm{\WB_t-\WB_\ast}^2\norm{\vb_\ast}^2-\frac{1}{2}\left(\frac{\sigma_m}{2}\norm{\vb_t-\vb_\ast}^2+\frac{2}{\sigma_m}\eta_v^2\right)\\
	&=\frac{\sigma_m}{4} \norm{\vb_t-\vb_\ast}^2-\frac{L^2}{2\sigma_m}\norm{\WB_t-\WB_\ast}^2\norm{\vb_\ast}^2-\frac{1}{\sigma_m}\eta_v^2.
	\end{align}
	The second inequality is due to the definition that $\sigma_m$ is the smallest non-negative eigenvalue of the matrix $\phi_{t,t}$ and \eqref{phi_L}.
	Therefore, we have 
	\begin{align}\label{vt_intmd}
	&\norm{\vb_{t+1}-\vb_\ast}^2 = \norm{\vb_t-\alpha\gb_t^v-\vb_\ast}^2=\norm{\vb_t-\vb_\ast}^2-2\alpha(\vb_t-\vb_\ast)^\top\gb_t^v+\alpha^2\norm{\gb_t^v}^2\nn\\
	&\le \norm{\vb_t-\vb_\ast}^2 - \frac{\alpha\sigma_m}{2} \norm{\vb_t-\vb_\ast}^2+\frac{\alpha L^2}{\sigma_m}\norm{\WB_t-\WB_\ast}^2\norm{\vb_\ast}^2+\frac{2\alpha}{\sigma_m}\eta_v^2 +\alpha^2\norm{\gb_t^v}^2.
	\end{align}
	Now, we can write
	\begin{align}
	\norm{\gb_t^v} &\le \norm{\bar \gb_t^v} +\eta_v \le \norm{\phi_{t,t}(\vb_t-\vb_\ast)+\left(\phi_{t,t} -\phi_{t,\ast} \right)\vb_\ast}+\eta_v\\
	&\le \norm{\phi_{t,t}(\vb_t-\vb_\ast)}+\norm{\left(\phi_{t,t} -\phi_{t,\ast} \right)\vb_\ast}+\eta_v\\
	&\le \sigma_M \norm{\vb_t-\vb_\ast} + L \norm{\WB_t-\WB_\ast}\norm{\vb_\ast}+\eta_v,
	\end{align} 
	where $\sigma_M$ is the largest non-negative eigenvalue of the matrix $\phi_{t,t}$.
	Hence, it results in
	\begin{align}
	\norm{\gb_t^v}^2\le 3\sigma^2_M \norm{\vb_t-\vb_\ast}^2 + 3L^2 \norm{\WB_t-\WB_\ast}^2\norm{\vb_\ast}^2+3\eta_v^2.
	\end{align}
	Plugging into the inequality \eqref{vt_intmd} provides
	\begin{align}
	\norm{\vb_{t+1}-\vb_\ast}^2 \le \left(1-\frac{\alpha\sigma_m}{2}+3\sigma^2_M\alpha^2\right)\norm{\vb_t-\vb_\ast}^2 +\left(\frac{\alpha L^2}{\sigma_m}+3L^2\alpha^2\right)\norm{\WB_t-\WB_\ast}^2\norm{\vb_\ast}^2+\left(\frac{2\alpha}{\sigma_m}+3\alpha^2\right)\eta_v^2.
	\end{align}
	
\end{proof}

\subsubsection{Proof of Lemma~\ref{node_level_lips}}
First, we can write
\begin{align}
&\norm{\GB^W(\WB,\vb) -\GB^W(\WB^\prime,\vb) } =\norm{ \frac{1}{n} \Xi^{-1}   \HB^\top \AB^\top (\DB^{-1})^\top  \left(\sigma (\DB^{-1}\AB\HB \WB)-\sigma (\DB^{-1}\AB\HB \WB^\prime) \right)\vb \vb^\top}\\
&\le  \frac{1}{n} \norm{\Xi^{-1} } \norm{\HB^\top \AB^\top (\DB^{-1})^\top}\norm{ \sigma (\DB^{-1}\AB\HB\WB)-\sigma (\DB^{-1}\AB\HB \WB^\prime) }\norm{\vb }^2\\
&\le \frac{1}{n} \norm{\Xi^{-1} }   \norm{\HB^\top \AB^\top (\DB^{-1})^\top}\norm{ \DB^{-1}\AB\HB \WB-\DB^{-1}\AB\HB \WB^\prime }\norm{\vb }^2.
\end{align}

Then, using the result from Lemma~\ref{node_lip} with probability at least $1-\delta$ if $n\ge \frac{8\ln \frac{1}{\delta}}{d d_{\min}}$
\begin{align}
\norm{\GB^W(\WB,\vb) -\GB^W(\WB^\prime,\vb) } &\le D_0^2  \norm{\Xi^{-1}}  \frac{1}{n}\norm{ \DB^{-1}\AB\HB}^2\norm{\WB- \WB^\prime} \\
&\le D_0^2  \norm{\Xi^{-1}} \norm{\WB- \WB^\prime}(d+4d\bar{d}) .
\end{align}
Thus, we have the desired result as
\begin{align}
\frac{\norm{\GB^W(\WB,\vb) -\GB^W(\WB^\prime,\vb) }}{\norm{\WB-\WB^\prime}} \le D_0^2  \norm{\Xi^{-1}} (d+4d\bar{d}) .
\end{align}

Next, we can have with probability at least $1-\delta$ if $n\ge \frac{8\ln \frac{1}{\delta}}{d d_{\min}}$
\begin{align}
&\norm{\GB^W(\WB,\vb) -\GB^W(\WB,\vb^\prime) } =\norm{ \frac{1}{n} \Xi^{-1}   \HB^\top \AB^\top (\DB^{-1})^\top  \sigma (\DB^{-1}\AB\HB \WB)\left(\vb \vb^\top-\vb^\prime \vb^{\prime\top }\right)}\\
&\le  L_\sigma \frac{1}{n} \norm{\Xi^{-1}}  \norm{\HB^\top \AB^\top (\DB^{-1})^\top}\norm{ \DB^{-1}\AB\HB \WB }\norm{\vb \vb^\top-\vb^\prime \vb^{\prime\top }}\\
&\le \norm{\Xi^{-1}}(d+4d\bar{d})L_\sigma\norm{\vb \vb^\top-\vb^\prime \vb^{\prime\top }} \\
& \le \norm{\Xi^{-1}}(d+4d\bar{d})L_\sigma \left(\norm{(\vb+\vb^\prime)(\vb-\vb^\prime)^\top}+\norm{\vb^\prime\vb^\top-\vb\vb^{\prime\top}}\right) \\
&  \le \norm{\Xi^{-1}}(d+4d\bar{d})L_\sigma \left(\norm{(\vb+\vb^\prime)(\vb-\vb^\prime)^\top}+\norm{(\vb^{\prime}-\vb)\vb^\top-\vb(\vb^{\prime}-\vb)^\top}\right)\\
&\le \norm{\Xi^{-1}}(d+4d\bar{d})L_\sigma \left(\norm{(\vb+\vb^\prime)}\norm{\vb-\vb^\prime}+2\norm{\vb^{\prime}-\vb}\norm{\vb}\right)\\
& \le 4D_0\norm{\Xi^{-1}}(d+4d\bar{d})L_\sigma\norm{\vb-\vb^\prime}.
\end{align}
Thus, we obtain
\begin{align}
\frac{\norm{\GB^W(\WB,\vb) -\GB^W(\WB,\vb^\prime) }}{\norm{\vb-\vb^\prime}} \le 4D_0\norm{\Xi^{-1}}(d+4d\bar{d})L_\sigma.
\end{align}

To show further results, we first write
\begin{align}
&g_1(\WB)= -\frac{1}{n} \sigma(\WB^\top \HB^\top \AB^\top (\DB^{-1})^\top)  \sigma (\DB^{-1}\AB\HB \WB_\ast)\vb_\ast,\\
&g_2(\WB)	= -\frac{1}{n}  \sigma(\WB^\top \HB^\top \AB^\top (\DB^{-1})^\top)  \eb,
\\
&g_3(\WB)= \frac{1}{n}  \sigma(\WB^\top \HB^\top \AB^\top (\DB^{-1})^\top) \sigma (\DB^{-1}\AB\HB \WB)\vb.
\end{align}
Then, we can express
\begin{align}
\norm{\gb^v(\WB,\vb)-\gb^v(\WB^\prime,\vb)}=\norm{g_1(\WB) -g_1(\WB^\prime) +g_2(\WB) -g_2(\WB^\prime)+ g_3(\WB)- g_3(\WB^\prime) }.
\end{align}
Additionally, we define
\begin{align}
g_3^\prime(\WB) = \sigma(\WB^\top \HB^\top \AB^\top (\DB^{-1})^\top) \sigma (\DB^{-1}\AB\HB \WB).
\end{align}
Therefore, we obtain
\begin{align}
&\frac{\norm{g^\prime_3(\WB) -g^\prime_3(\WB^\prime)} }{\norm{\WB-\WB^\prime}} \le \frac{\norm{g^\prime_3(\WB) -g^\prime_3(\WB^\prime)} }{\norm{\sigma (\DB^{-1}\AB\HB \WB)-\sigma (\DB^{-1}\AB\HB \WB^\prime)}}\frac{\norm{ \sigma (\DB^{-1}\AB\HB \WB)-\sigma (\DB^{-1}\AB\HB \WB^\prime)}}{\norm{\WB-\WB^\prime}} \\
&\le2 \left(\norm{ \sigma (\DB^{-1}\AB\HB \WB)}+\norm{\sigma (\DB^{-1}\AB\HB \WB^\prime)}\right) {\norm{ \DB^{-1}\AB\HB }} \\
&\le 4L_\sigma \norm{ \DB^{-1}\AB\HB  }\norm{ \DB^{-1}\AB\HB  }.
\end{align}
Thus, it results in 
\begin{align}
\frac{\norm{g_1(\WB) -g_1(\WB^\prime)} }{\norm{\WB-\WB^\prime}}  \le L_\sigma \frac{1}{n} \norm{\HB^\top \AB^\top (\DB^{-1})^\top}\norm{ \DB^{-1}\AB\HB }\norm{\vb_\ast},
\end{align}
\begin{align}
\frac{\norm{g_3(\WB) -g_3(\WB^\prime)} }{\norm{\WB-\WB^\prime}} \le \frac{\norm{\vb}}{n}  4L_\sigma \norm{ \DB^{-1}\AB\HB  }\norm{ \DB^{-1}\AB\HB  }.
\end{align}

Therefore,  if $n\ge \frac{8\ln \frac{1}{\delta}}{d d_{\min}}$ with probability at least $1-\delta$
\begin{align}
\frac{\norm{g_3(\WB) -g_3(\WB^\prime)}+\norm{g_1(\WB) -g_1(\WB^\prime)} }{\norm{\WB-\WB^\prime}} \le 5(d+4d\bar{d})D_0L_\sigma .
\end{align}
We also have
\begin{align}
&\frac{\norm{g_2(\WB) -g_2(\WB^\prime)} }{\norm{\WB-\WB^\prime}} 	= \frac{\norm{\frac{1}{n}  \left[\sigma(\WB^\top \HB^\top \AB^\top (\DB^{-1})^\top)-\sigma(\WB^{\prime\top} \HB^\top \AB^\top (\DB^{-1})^\top)\right]\eb} }{\norm{\WB-\WB^\prime}}\\
& \le \frac{\frac{1}{n} \norm{ (\WB^\top-\WB^{\prime\top}) \HB^\top \AB^\top (\DB^{-1})^\top}\norm{\eb} }{\norm{\WB^\top-\WB^{\prime\top}}}\\
&\le {\frac{1}{n} \left(  \norm{ \HB^\top \AB^\top (\DB^{-1})^\top}  \norm{\eb} \right) }\\
& \le {\frac{1}{2n} \left(  \norm{ \HB^\top \AB^\top (\DB^{-1})^\top}^2+   \norm{\eb}^2 \right) }.
\end{align}

According to Lemma~\ref{node_lip}, if $n\ge \frac{8\ln \frac{2}{\delta}}{d d_{\min}}$, with a probability at least $1-\frac{\delta}{2}$ 
\begin{align}
\frac{1}{2n} \norm{ \HB^\top \AB^\top (\DB^{-1})^\top}^2 \le \frac{d+4d\bar{d}}{2},
\end{align}
and with a probability at least $1-\frac{\delta}{2}$ 
\begin{align}
\frac{1}{2n}  \norm{\eb}^2  \le \nu^2.
\end{align}
By union bound, with a probability at least $1-{\delta}$ 
\begin{align}
 {\frac{1}{2n} \left(  \norm{ \HB^\top \AB^\top (\DB^{-1})^\top}^2+   \norm{\eb}^2 \right) } \le  \frac{d+4d\bar{d}}{2}+ \nu^2.
\end{align}
Thus,  if $n\ge \frac{8\ln \frac{2}{\delta}}{d d_{\min}}$, with probability at least $1-{\delta}$ .
\begin{align}
\frac{\norm{g_2(\WB) -g_2(\WB^\prime)} }{\norm{\WB-\WB^\prime}} \le   \frac{d+4d\bar{d}}{2}+ \nu^2
\end{align}
Therefore, if $n\ge \frac{8\ln \frac{2}{\delta}}{d d_{\min}}$, with probability at least $1-\delta$
\begin{align}
\frac{\norm{\gb^v(\WB,\vb)-\gb^v(\WB^\prime,\vb)}}{\norm{\WB-\WB^\prime}} \le 5(d+4d\bar{d})D_0L_\sigma + \frac{d+4d\bar{d}}{2}+ \nu^2.
\end{align}

At last, we can write
\begin{align}
&\frac{\norm{\gb^v(\WB,\vb) -\gb^v(\WB,\vb^\prime) }}{\norm{\vb-\vb^\prime}}	= \frac{g_3(\WB,\vb)-g_3(\WB,\vb^\prime)}{\norm{\vb-\vb^\prime}} 
&\le \frac{L_\sigma^2}{n} \norm{ \DB^{-1}\AB\HB  }\norm{ \DB^{-1}\AB\HB  }.
\end{align}
Thus, if $n\ge \frac{8\ln \frac{1}{\delta}}{d d_{\min}}$, with probability at least $1-\delta$,
\begin{align}
\frac{\norm{\gb^v(\WB,\vb) -\gb^v(\WB,\vb^\prime) }}{\norm{\vb-\vb^\prime}} \le {L_\sigma^2}(d+4d\bar{d}).
\end{align}

\subsubsection{Proof of Lemma~\ref{node_level_eta}}
\begin{proof}
	$\GB^W-\bar\GB^W$ is a centered sub-Gaussian random matrix, and with straightforward calculation, we have
	\begin{align}
	\GB^W = \frac{1}{n} {\Xi^{-1}}U,
	\end{align}
	where
	$
	U=\sum_{i=1}^n(U_1+U_2+U_3)\vb^\top
	$
	and
	\begin{align}
	&U_1 =\frac{\sum_{j\in \NC_i}\HB^\top_j}{d_i} \sigma\left(\frac{\sum_{j\in \NC_i}\HB_j\WB_{\ast}}{d_i}\right)\vb_\ast,\\
	&U_2 = \frac{\sum_{j\in \NC_i}\HB^\top_j}{d_i} \epsilon_i,\\
	&U_3 = - \frac{\sum_{j\in \NC_i}\HB^\top_j}{d_i} \sigma\left(\frac{\sum_{j\in \NC_i}\HB_j\WB}{d_i}\right)\vb,
	\end{align}
	where $\NC_i$ is the set containing node $i$ and all its neighboring nodes, and the degree $d_i$ equals the cardinality of $\NC_i$.

	According to Lemma B.8 in \cite{cao2019tight}, we have $\normpsi{\ab^\top( U_1+U_3)} \le c_1D_0\frac{d}{d_i}(1+|\sigma(0)|)$ and $\normpsi{\ab^\top U_2} \le c_2\sqrt{\frac{d}{d_i}}\nu$  for some absolute constants $c_1$ and $c_2$. Thus, by Lemmas D. 3 and D. 4 in \cite{yi2015regularized}, we have
	\begin{align}
	\normpsia{\ab^\top (U_1+U_2+U_3)\vb^\top \bb} \le c_3 D_0 \left(D_0\frac{d}{d_i}(1+|\sigma(0)|)+\sqrt{\frac{d}{d_i}}\nu\right) 
	\end{align}
 for some absolute constant $c_3$, 
	where $\ab \in \mathcal{N}_1$ and $ \mathcal{N}_1 = \mathcal{N}(S^{d-1},1/2)$ is a 1/2-net covering $S^{d-1}$, and $\bb \in \mathcal{N}_2$ and $ \mathcal{N}_2 = \mathcal{N}(S^{d_{out}-1},1/2)$ is a 1/2-net covering $S^{d_{out}-1}$. Note that $|\mathcal{N}_1| \le 5^d$ and $|\mathcal{N}_2| \le 5^{d_{out}}$. 
	According to Proposition 5.16 in \cite{tropp2015introduction},
	\begin{align}
	P\left(\left| \ab^\top (\GB^W-\bar\GB^W) \bb\right|  \ge \frac{1}{c}\sqrt{{\frac{\log n}{n}}} T\right) \le 2\exp\left(-\frac{\sqrt{{n\log n}}T}{K}\right),
	\end{align}
	where $K=T=\norm{\Xi^{-1}}D_0 \left(D_0\frac{d}{d_{\min}}(1+|\sigma(0)|)+\sqrt{\frac{d}{d_{\min}}}\nu\right) $ and $d_{\min}$ is the smallest degree of the nodes in the graph. Then if $\sqrt{n\log n}\ge \frac{3}{2} \log \frac{2}{\delta}$, with probability at least $1-\delta$ 
	\begin{align}
	\left| \ab^\top (\GB^W-\bar\GB^W) \bb\right|  \le \frac{1}{c}\sqrt{{\frac{\log n}{n}}} \norm{\Xi^{-1}}D_0 \left(D_0\frac{d}{d_{\min}}(1+|\sigma(0)|)+\sqrt{\frac{d}{d_{\min}}}\nu\right) .
	\end{align}
	By Lemma 5.3 in \cite{tropp2015introduction},  if $\sqrt{n\log n}\ge \frac{3}{2} \log \frac{2}{\delta}$, with probability at least $1-\delta$ 
	\begin{align}
	\norm{\GB^W-\bar\GB^W } \le \frac{4}{c}\sqrt{{\frac{\log n}{n}}} \norm{\Xi^{-1}}D_0 \left(D_0\frac{d}{d_{\min}}(1+|\sigma(0)|)+\sqrt{\frac{d}{d_{\min}}}\nu\right) ,
	\end{align}
	for some absolute constant $c$.

	Similarly, 	$\gb^v-\bar\gb^v$ is a centered sub-Gaussian random matrix, and with straightforward calculation, we have
	\begin{align}
	\gb^v = \frac{1}{n}\sum_{i=1}^n(U_1+U_2+U_3),
	\end{align}
	where
	\begin{align}
	&U_1 =\sigma\left(\frac{\sum_{j\in \NC_i}\WB^\top \HB_j^\top}{d_i}\right) \sigma\left(\frac{\sum_{j\in \NC_i}\HB_j\WB_{\ast}}{d_i}\right)\vb_\ast,\\
	&U_2 = \sigma\left(\frac{\sum_{j\in \NC_i}\WB^\top \HB_j^\top}{d_i}\right)\epsilon_i,\\
	&U_3 = - \sigma\left(\frac{\sum_{j\in \NC_i}\WB^\top \HB_j^\top}{d_i}\right)\sigma\left(\frac{\sum_{j\in \NC_i}\HB_j\WB}{d_i}\right)\vb,
	\end{align}
	where $\NC_i$ is the set containing node $i$ and all its neighboring nodes, and the degree $d_i$ equals the cardinality of $\NC_i$.

	According to Lemma B.8 in \cite{cao2019tight}, we have $\normpsi{\ab^\top( U_1+U_3)} \le c_1D_0\frac{d}{d_i}(1+|\sigma(0)|)^2$ and $\normpsi{\ab^\top U_2} \le c_2\sqrt{\frac{d}{d_i}}(1+|\sigma(0)|)\nu$  for some absolute constants $c_1$ and $c_2$. Thus, by Lemmas D. 3 and D. 4 in \cite{yi2015regularized}, we have
	\begin{align}
	\normpsia{\ab^\top (U_1+U_2+U_3)} \le c_3  \left(D_0\frac{d}{d_i}(1+|\sigma(0)|)^2+\sqrt{\frac{d}{d_i}}(1+|\sigma(0)|)\nu\right) 
	\end{align}
 for some absolute constant $c_3$, 
	where $\ab \in \mathcal{N}_1$ and $ \mathcal{N}_1 = \mathcal{N}(S^{d-1},1/2)$ is a 1/2-net covering $S^{d-1}$. Note that $|\mathcal{N}_1| \le 5^d$ by Lemma 5.2 in \cite{tropp2015introduction}. 
	According to Proposition 5.16 in \cite{tropp2015introduction},
	\begin{align}
	P\left(\left| \ab^\top (\gb^v-\bar\gb^v) \right|  \ge \frac{1}{c}\sqrt{{\frac{\log n}{n}}} T\right) \le 2\exp\left(-\frac{\sqrt{{n\log n}}T}{K}\right),
	\end{align}
	where $K=T= \left(D_0\frac{d}{d_{\min}}(1+|\sigma(0)|)^2+\sqrt{\frac{d}{d_{\min}}}(1+|\sigma(0)|)\nu\right) $ and $d_{\min}$ is the smallest degree of the nodes in the graph. Then if $\sqrt{n\log n}\ge \frac{3}{2}\log \frac{2}{\delta}$, with probability at least $1-\delta$ 
	\begin{align}
	\left| \ab^\top (\gb^v-\bar\gb^v) \right|  \le \frac{1}{c}\sqrt{{\frac{\log n}{n}}}  \left(D_0\frac{d}{d_{\min}}(1+|\sigma(0)|)^2+\sqrt{\frac{d}{d_{\min}}}(1+|\sigma(0)|)\nu\right) ,
	\end{align}
	By Lemma 5.3 in \cite{tropp2015introduction},  if $\sqrt{n\log n}\ge \frac{3}{2} \log \frac{2}{\delta}$, with probability at least $1-\delta$ 
	\begin{align}
	\norm{(\gb^v-\bar\gb^v) } \le \frac{2}{c}\sqrt{{\frac{\log n}{n}}} \left(D_0\frac{d}{d_{\min}}(1+|\sigma(0)|)^2+\sqrt{\frac{d}{d_{\min}}}(1+|\sigma(0)|)\nu\right) ,
	\end{align}
	for some absolute constant $c$.
\end{proof}

\subsubsection*{Proof of Lemma~\ref{graph_level_lips}}
\begin{proof}
	First, we can write
	\begin{align}
	&\norm{\GB^W(\WB,\vb) -\GB^W(\WB^\prime,\vb) } =\norm{ \frac{1}{n} \norm{\Xi^{-1}} \sum_{j=1}^{n}  \HB_{j}^\top \AB_{j}^\top (\DB_{j}^{-1})^\top \ab_j^\top  \ab_j \left(\sigma (\DB_j^{-1}\AB_j\HB_j \WB)-\sigma (\DB_j^{-1}\AB_j\HB_j \WB^\prime) \right)\vb \vb^\top}\\
	&\le  \frac{1}{n} \norm{\Xi^{-1}} \sum_{j=1}^{n} \norm{\HB_{j}^\top \AB_{j}^\top (\DB_{j}^{-1})^\top}\norm{ \ab_j}^2\norm{ \sigma (\DB_j^{-1}\AB_j\HB_j \WB)-\sigma (\DB_j^{-1}\AB_j\HB_j \WB^\prime) }\norm{\vb }^2\\
	&\le \frac{1}{n} \norm{\Xi^{-1}} \sum_{j=1}^{n} \norm{\HB_{j}^\top \AB_{j}^\top (\DB_{j}^{-1})^\top}\norm{ \ab_j}^2\norm{ \DB_j^{-1}\AB_j\HB_j \WB-\DB_j^{-1}\AB_j\HB_j \WB^\prime }\norm{\vb }^2.
	\end{align}
	According to Lemma~\ref{lip}, if $\sum_{j=1}^{n}n_j \ge\frac{8\ln\frac{1}{\delta}}{d} $ with probability at least $1-\delta$
	\begin{align}
	\norm{\GB^W(\WB,\vb) -\GB^W(\WB^\prime,\vb) }& \le D_0^2  \norm{\Xi^{-1}} n_{\max}\frac{1}{n}\sum_{j=1}^{n} \norm{\HB_j}^2\norm{\WB- \WB^\prime} \\
	&\le D_0^2  \norm{\Xi^{-1}} n_{\max}\frac{1}{n}2d\sum_{j=1}^{n}n_j\norm{\WB- \WB^\prime} \\
	& \le 2dD_0^2  \norm{\Xi^{-1}} n^2_{\max}\norm{\WB- \WB^\prime}.
	\end{align}
	Thus, we have the desired result as
	\begin{align}
	\frac{\norm{\GB^W(\WB,\vb) -\GB^W(\WB^\prime,\vb) }}{\norm{\WB-\WB^\prime}} \le 2dD_0^2  \norm{\Xi^{-1}} n^2_{\max} .
	\end{align}
	
	Similarly, we have  with probability at least $1-\delta$ if $\sum_{j=1}^{n}n_j \ge\frac{8\ln\frac{1}{\delta}}{d} $
	\begin{align}
	&\norm{\GB^W(\WB,\vb) -\GB^W(\WB,\vb^\prime) } =\norm{ \frac{1}{n} \norm{\Xi^{-1}} \sum_{j=1}^{n}  \HB_{j}^\top \AB_{j}^\top (\DB_{j}^{-1})^\top \ab_j^\top  \ab_j \sigma (\DB_j^{-1}\AB_j\HB_j \WB)\left(\vb \vb^\top-\vb^\prime \vb^{\prime\top }\right)}\\
	&\le  L_\sigma \frac{1}{n} \norm{\Xi^{-1}} \sum_{j=1}^{n} \norm{\HB_{j}^\top \AB_{j}^\top (\DB_{j}^{-1})^\top}\norm{ \ab_j}^2\norm{ \DB_j^{-1}\AB_j\HB_j \WB }\norm{\vb \vb^\top-\vb^\prime \vb^{\prime\top }}\\
	&\le 2d\norm{\Xi^{-1}}n^2_{\max}L_\sigma\norm{\vb \vb^\top-\vb^\prime \vb^{\prime\top }} \\
	& \le 2d\norm{\Xi^{-1}}n^2_{\max}L_\sigma \left(\norm{(\vb+\vb^\prime)(\vb-\vb^\prime)^\top}+\norm{\vb^\prime\vb^\top-\vb\vb^{\prime\top}}\right) \\
	&  \le 2d\norm{\Xi^{-1}}n^2_{\max}L_\sigma \left(\norm{(\vb+\vb^\prime)(\vb-\vb^\prime)^\top}+\norm{(\vb^{\prime}-\vb)\vb^\top-\vb(\vb^{\prime}-\vb)^\top}\right)\\
	&\le 2d\norm{\Xi^{-1}}n^2_{\max}L_\sigma \left(\norm{(\vb+\vb^\prime)}\norm{\vb-\vb^\prime}+2\norm{\vb^{\prime}-\vb}\norm{\vb}\right)\\
	& \le 8D_0d\norm{\Xi^{-1}}n^2_{\max}L_\sigma\norm{\vb-\vb^\prime}.
	\end{align}
	Thus, we obtain
	\begin{align}
	\frac{\norm{\GB^W(\WB,\vb) -\GB^W(\WB,\vb^\prime) }}{\norm{\vb-\vb^\prime}} \le 8D_0d\norm{\Xi^{-1}}n^2_{\max}L_\sigma.
	\end{align}
	
	To show further results, we first write
	\begin{align}
	&g_1(\WB)= -\frac{1}{n} \sum_{j=1}^{n} \sigma(\WB^\top \HB_{j}^\top \AB_{j}^\top (\DB_{j}^{-1})^\top) \ab_j^\top  \ab_j \sigma (\DB_j^{-1}\AB_j\HB_j \WB_\ast)\vb_\ast,\\
	&g_2(\WB)	= -\frac{1}{n} \sum_{j=1}^{n} \sigma(\WB^\top \HB_{j}^\top \AB_{j}^\top (\DB_{j}^{-1})^\top) \ab_j^\top  \ab_j e_j,
	\\
	&g_3(\WB)= \frac{1}{n} \sum_{j=1}^{n} \sigma(\WB^\top \HB_{j}^\top \AB_{j}^\top (\DB_{j}^{-1})^\top) \ab_j^\top  \ab_j \sigma (\DB_j^{-1}\AB_j\HB_j \WB)\vb.
	\end{align}
	Then, we can express
	\begin{align}
	\norm{\gb^v(\WB,\vb)-\gb^v(\WB^\prime,\vb)}=\norm{g_1(\WB) -g_1(\WB^\prime) +g_2(\WB) -g_2(\WB^\prime)+ g_3(\WB)- g_3(\WB^\prime) }.
	\end{align}
	Additionally, we define
	\begin{align}
	g_3^\prime(\WB) = \sigma(\WB^\top \HB_{j}^\top \AB_{j}^\top (\DB_{j}^{-1})^\top) \ab_j^\top  \ab_j \sigma (\DB_j^{-1}\AB_j\HB_j \WB).
	\end{align}
	Therefore, we obtain
	\begin{align}
	&\frac{\norm{g^\prime_3(\WB) -g^\prime_3(\WB^\prime)} }{\norm{\WB-\WB^\prime}} \le \frac{\norm{g^\prime_3(\WB) -g^\prime_3(\WB^\prime)} }{\norm{\ab_j \sigma (\DB_j^{-1}\AB_j\HB_j \WB)-\ab_j \sigma (\DB_j^{-1}\AB_j\HB_j \WB^\prime)}}\frac{\norm{\ab_j \sigma (\DB_j^{-1}\AB_j\HB_j \WB)-\ab_j \sigma (\DB_j^{-1}\AB_j\HB_j \WB^\prime)}}{\norm{\WB-\WB^\prime}} \\
	&\le2 \left(\norm{\ab_j \sigma (\DB_j^{-1}\AB_j\HB_j \WB)}+\norm{\ab_j \sigma (\DB_j^{-1}\AB_j\HB_j \WB^\prime)}\right) {\norm{\ab_j  \DB_j^{-1}\AB_j\HB_j }} \\
	&\le 4L_\sigma \norm{ \DB_j^{-1}\AB_j\HB_j  }\norm{ \ab_j}^2\norm{ \DB_j^{-1}\AB_j\HB_j  }.
	\end{align}
	Thus, it results in
	\begin{align}
	\frac{\norm{g_1(\WB) -g_1(\WB^\prime)} }{\norm{\WB-\WB^\prime}}  \le L_\sigma \frac{1}{n} \sum_{j=1}^{n} \norm{\HB_{j}^\top \AB_{j}^\top (\DB_{j}^{-1})^\top}\norm{ \ab_j}^2\norm{ \DB_j^{-1}\AB_j\HB_j }\norm{\vb_\ast},
	\end{align}
	\begin{align}
	\frac{\norm{g_3(\WB) -g_3(\WB^\prime)} }{\norm{\WB-\WB^\prime}} \le \frac{\norm{\vb}}{n} \sum_{j=1}^{n} 4L_\sigma \norm{ \DB_j^{-1}\AB_j\HB_j  }\norm{ \ab_j}^2\norm{ \DB_j^{-1}\AB_j\HB_j  }.
	\end{align}

	Therefore,  if $\sum_{j=1}^{n}n_j \ge\frac{8\ln\frac{1}{\delta}}{d} $  with probability $1-\delta$
	\begin{align}
	\frac{\norm{g_3(\WB) -g_3(\WB^\prime)}+\norm{g_1(\WB) -g_1(\WB^\prime)} }{\norm{\WB-\WB^\prime}} \le 10dD_0L_\sigma n_{\max}^2.
	\end{align}
	We also have
	\begin{align}
	&\frac{\norm{g_2(\WB) -g_2(\WB^\prime)} }{\norm{\WB-\WB^\prime}} 	= \frac{\norm{\frac{1}{n} \sum_{j=1}^{n} \left[\sigma(\WB^\top \HB_{j}^\top \AB_{j}^\top (\DB_{j}^{-1})^\top)-\sigma(\WB^{\prime\top} \HB_{j}^\top \AB_{j}^\top (\DB_{j}^{-1})^\top)\right] \ab_j^\top  \ab_j e_j} }{\norm{\WB-\WB^\prime}}\\
	& \le \frac{\frac{1}{n} \sum_{j=1}^{n}\norm{ (\WB^\top-\WB^{\prime\top}) \HB_{j}^\top \AB_{j}^\top (\DB_{j}^{-1})^\top}| e_j| }{\norm{\WB^\top-\WB^{\prime\top}}}\\
	&\le {\frac{1}{n} \sum_{j=1}^{n}\norm{ \HB_{j}^\top \AB_{j}^\top (\DB_{j}^{-1})^\top}| e_j| }\\
	&\le {\frac{1}{n} \sum_{j=1}^{n}\norm{ \HB_{j}}\norm{ \AB_{j}^\top (\DB_{j}^{-1})^\top}| e_j| } \\
	& \le {\frac{1}{n} \sum_{j=1}^{n}\sqrt{n_{j}}\norm{ \HB_{j}}| e_j| } \\
	& \le {\frac{\sqrt{n_{\max}}}{2n} \sum_{j=1}^{n}\norm{ \HB_{j}}^2+e_j^2 }.
	\end{align}
	
	According to Lemma~\ref{lip}, if $\sum_{j=1}^{n}n_j \ge \frac{8\ln\frac{2}{\delta}}{d}$, with a probability at least $1-\frac{\delta}{2}$ 
	\begin{align}
	\sum_{j=1}^{n}\norm{\HB_j}^2 \le 2d\sum_{j=1}^{n}n_j  ,
	\end{align}
	and with a probability at least $1-\frac{\delta}{2}$ 
	\begin{align}
	\sum_{j=1}^{n}e_j^2 \le 2n\nu^2.
	\end{align}
	By union bound, with probability at least $1-{\delta}$ 
	\begin{align}
	\sum_{j=1}^{n}\norm{ \HB_{j}}^2+e_j^2 \le 2d\sum_{j=1}^{n}n_j + 2n\nu^2.
	\end{align}
	Thus, if $\sum_{j=1}^{n}n_j \ge \frac{8\ln\frac{2}{\delta}}{d}$, with probability at least $1-{\delta}$ 
	\begin{align}
	\frac{\norm{g_2(\WB) -g_2(\WB^\prime)} }{\norm{\WB-\WB^\prime}} \le  {\frac{\sqrt{n_{\max}}}{2n} \left(2d\sum_{j=1}^{n}n_j + 2n\nu^2\right)} \le {d\sqrt{n^3_{\max}}}+{\sqrt{n_{\max}}}\nu^2.
	\end{align}
	Therefore, if $\sum_{j=1}^{n}n_j \ge \frac{8\ln\frac{2}{\delta}}{d}$, with probability at least $1-\delta$
	\begin{align}
	\frac{\norm{\gb^v(\WB,\vb)-\gb^v(\WB^\prime,\vb)}}{\norm{\WB-\WB^\prime}} \le 10dD_0L_\sigma n_{\max}^2+{d\sqrt{n^3_{\max}}}+{\sqrt{n_{\max}}}\nu^2.
	\end{align}
	
	At last, we can write
	\begin{align}
	\frac{\norm{\gb^v(\WB,\vb) -\gb^v(\WB,\vb^\prime) }}{\norm{\vb-\vb^\prime}}	&= \frac{g_3(\WB,\vb)-g_3(\WB,\vb^\prime)}{\norm{\vb-\vb^\prime}} \\
	&\le \frac{L_\sigma^2}{n} \sum_{j=1}^{n}  \norm{ \DB_j^{-1}\AB_j\HB_j  }\norm{ \ab_j}^2\norm{ \DB_j^{-1}\AB_j\HB_j  }.
	\end{align}
	Thus, if $\sum_{j=1}^{n}n_j \ge \frac{8\ln\frac{1}{\delta}}{d}$, with probability at least $1-\delta$,
	\begin{align}
	\frac{\norm{\gb^v(\WB,\vb) -\gb^v(\WB,\vb^\prime) }}{\norm{\vb-\vb^\prime}} \le 2d{L_\sigma^2}n_{\max}^2.
	\end{align}
\end{proof}

\subsubsection*{Proof of Lemma~\ref{graph_level_eta}}
\begin{proof}
	Define $\sum_{j=1}^n\mathit{G}_j=n(\GB^W-\bar\GB^W)$, where $\mathcal{G}_j$ is a centered sub-Gaussian random matrix. According to Lemma B.8 in \cite{cao2019tight}, we have $\normpsi{\sigma (\DB_j^{-1}\AB_j\HB_j \WB)} \le c_1\sqrt{n_j}(1+|\sigma(0)|)$ and $\normpsi{ \ab^\top\DB_j^{-1}\AB_j\HB_j \WB} \le c_2\sqrt{n_j}$ for some absolute constants $c_1$ and $c_2$. Thus, by Lemmas D. 3 and D. 4 in \cite{yi2015regularized}, we have
	\begin{align}
	\normpsia{\ab^\top\mathit{G}_j\bb} &\le c_3\norm{\Xi^{-1}} D_0\sqrt{n_j} \left(\sqrt{n_j}(1+|\sigma(0)|)(\norm{\vb_\ast}+D_0)+\nu\right) \\
	&\le c_4\norm{\Xi^{-1}}D_0 \sqrt{n_j} \left(\sqrt{n_j}(1+|\sigma(0)|)D_0+\nu\right)
	\end{align}
 for some absolute constants $c_3$ and $c_4$, 
	where $\ab \in \mathcal{N}_1$ and $ \mathcal{N}_1 = \mathcal{N}(S^{d-1},1/2)$ is a 1/2-net covering $S^{d-1}$, and $\bb \in \mathcal{N}_2$ and $ \mathcal{N}_2 = \mathcal{N}(S^{d_{out}-1},1/2)$ is a 1/2-net covering $S^{d_{out}-1}$. Note that $|\mathcal{N}_1| \le 5^d$ and $|\mathcal{N}_2| \le 5^{d_{out}}$. 
	According to Proposition 5.16 in \cite{tropp2015introduction},
	\begin{align}
	P\left(\left| \sum_{j=1}^{n}\ab^\top G_j \bb\right|  \ge \frac{1}{c}\sqrt{{n\log n}} T\right) \le 2\exp\left(-\frac{\sqrt{{n\log n}}T}{K}\right),
	\end{align}
	where $K=T=c_4\norm{\Xi^{-1}}D_0 \sqrt{n_{\max}} \left(\sqrt{n_{\max}}(1+|\sigma(0)|)D_0+\nu\right)$. Then if $\sqrt{n\log n}\ge\frac{3}{2} \log \frac{2}{\delta}$, with probability at least $1-\delta$ 
	\begin{align}
	\left| \sum_{j=1}^{n}\ab^\top G_i\bb\right|  \le \frac{1}{c}\sqrt{{n\log n}} \norm{\Xi^{-1}}D_0 \sqrt{n_{\max}} \left(\sqrt{n_{\max}}(1+|\sigma(0)|)D_0+\nu\right).
	\end{align}
	By Lemma 5.3 in \cite{tropp2015introduction},  if $\sqrt{n\log n}\ge \frac{3}{2} \log \frac{2}{\delta}$, with probability at least $1-\delta$ 
	\begin{align}
	\norm{\GB^W-\bar\GB^W } \le \frac{4}{c}\sqrt{\frac{\log n}{n}} \norm{\Xi^{-1}} D_0\sqrt{n_{\max}}\left(\sqrt{n_{\max}}(1+|\sigma(0)|)D_0+\nu\right),
	\end{align}
	for some absolute constant $c$.

	Similarly, define $\sum_{j=1}^n g_j=n(\gb^v-\bar\gb^v)$, where $g_j$ is a centered sub-Gaussian random matrix. According to Lemma B.8 in \cite{cao2019tight}, we have $\normpsi{\cb^\top\sigma (\DB_j^{-1}\AB_j\HB_j \WB)} \le c_5\sqrt{n_j}(1+|\sigma(0)|)$ for some absolute constant $c_5$. Thus, by Lemmas D. 3 and D. 4 in \cite{yi2015regularized}, we have
	\begin{align}
	\normpsia{\cb^\top g_j} \le c_6 \sqrt{n_j}(1+|\sigma(0)|) \left(\sqrt{n_j}(1+|\sigma(0)|)D_0+\nu\right)
	\end{align}
 for some absolute constant $c_6$, 
	where $\cb \in \mathcal{N}_3$ and $ \mathcal{N}_3  = \mathcal{N}(S^{d_{out}-1},1/2)$ is a 1/2-net covering $S^{d_{out}-1}$. Note that $|\mathcal{N}_3| \le 5^{d_{out}}$. 
	According to Proposition 5.16 in \cite{tropp2015introduction},
	\begin{align}
	P\left(\left| \sum_{j=1}^{n}\cb^\top g_j\right|  \ge \frac{1}{c}\sqrt{{n\log n}} T\right) \le 2\exp\left(-\frac{\sqrt{{n\log n}}T}{K}\right),
	\end{align}
	where $K=T=c_6 	\sqrt{n_{\max}}(1+|\sigma(0)|) \left(	\sqrt{n_{\max}}(1+|\sigma(0)|)D_0+\nu\right)$. Then if $\sqrt{n\log n}\ge \frac{3}{2}\log \frac{2}{\delta}$, with probability at least $1-\delta$ 
	\begin{align}
	\left| \sum_{j=1}^{n}\cb^\top g_i\right|  \le \frac{1}{c}\sqrt{\frac{\log n}{n}} 	\sqrt{n_{\max}}(1+|\sigma(0)|) \left(	\sqrt{n_{\max}}(1+|\sigma(0)|)D_0+\nu\right).
	\end{align}
	By Lemma 5.3 in \cite{tropp2015introduction},  if $\sqrt{n\log n}\ge \frac{3}{2} \log \frac{2}{\delta}$, with probability at least $1-\delta$ 
	\begin{align}
	\norm{\gb^v-\bar\gb^v } \le \frac{2}{c}\sqrt{\frac{\log n}{n}} 	\sqrt{n_{\max}}(1+|\sigma(0)|) \left(	\sqrt{n_{\max}}(1+|\sigma(0)|)D_0+\nu\right),
	\end{align}
	for some absolute constant $c$.
	
\end{proof}

\subsubsection{Proof of Lemma~\ref{node_lip}}
\begin{proof}
	Note that each entry in the matrix $\DB^{-1}\AB\HB$ is a Gaussian variable with zero mean and variance $\frac{1}{d_i}$ and $i$ is the row (node) index of the entry, and $d_i$ is the degree of $i$th node. By Lemmas~\ref{node_SE_concentration} and \ref{node_SE}, if $n\ge \frac{8\ln \frac{1}{\delta}}{d d_{\min}}$ with $d_{\min}$ being the smallest node degree in the graph, then with probability at least $1-\delta$
	\begin{align} 
	\frac{1}{nd}   \norm{ \DB^{-1}\AB\HB}^2
	\le 1+4\frac{\sum_{i=1}^n\frac{1}{d_i}}{n} ,
	\end{align}
	leading to 
	\begin{align}
	\frac{1}{n}   \norm{ \DB^{-1}\AB\HB}^2
	\le d+4d\frac{\sum_{i=1}^n\frac{1}{d_i}}{n} = d+4d\bar{d} ,
	\end{align}
	where $\bar{d}=\frac{\sum_{i=1}^n\frac{1}{d_i}}{n}$.
\end{proof}

\subsubsection{Proof of Lemma~\ref{lip}}
\begin{proof}
	According to Lemma~\ref{chi_hoeffding}, if $\sum_{j=1}^{n}n_j \ge\frac{8\ln\frac{1}{\delta}}{d} $ with probability at least $1-\delta$
\begin{align}
 \frac{1}{n}  \sum_{j=1}^{n} \norm{\HB_{j}^\top \AB_{j}^\top (\DB_{j}^{-1})^\top}\norm{ \ab_j}^2\norm{ \DB_j^{-1}\AB_j\HB_j  }&\le  \frac{1}{n}  \sum_{j=1}^{n} \norm{ \DB_j^{-1}\AB_j}^2\norm{\HB_j}^2
\\
&\le  \frac{1}{n}  \sum_{j=1}^{n} n_j\norm{\HB_j}^2
\\&\le  n_{\max}\frac{1}{n}2d\sum_{j=1}^{n}n_j\\
&\le 2dn^2_{\max},
\end{align}
completing the proof.
\end{proof}

\subsubsection*{Proof of Theorem~\ref{convergence_thm} }
\begin{proof}
	Since we have
	\begin{align}
	\|\WB_{t+1} - \WB_\ast\|_2 - \frac{4\eta_W\left(1+\alpha\rho+\sqrt{1+\alpha\rho}\right)}{\rho}\le \frac{1}{\sqrt{1+\alpha\rho}}\left[	\|\WB_{t} - \WB_\ast\|_2-\frac{4\eta_W\left(1+\alpha\rho+\sqrt{1+\alpha\rho}\right)}{\rho}\right],
	\end{align}
	it directly follows
	\begin{align}
	\norm{\WB_{t} - \WB_\ast} &\le\left(\frac{1}{\sqrt{1+\alpha\rho}}\right)^t\left[\norm{\WB_0-\WB_\ast}-\frac{4\eta_W\left(1+\alpha\rho+\sqrt{1+\alpha\rho}\right)}{\rho}\right]+\frac{4\eta_W\left(1+\alpha\rho+\sqrt{1+\alpha\rho}\right)}{\rho}\\
	&\le \left(\frac{1}{\sqrt{1+\alpha\rho}}\right)^t\norm{\WB_0-\WB_\ast}+\frac{4\eta_W\left(1+\alpha\rho+\sqrt{1+\alpha\rho}\right)}{\rho}\\
	&	\le \left(\frac{1}{\sqrt{1+\alpha\rho}}\right)^t\norm{\WB_0-\WB_\ast}+\eta_W(6\alpha +\frac{8}{\rho}),
	\end{align}
	and we have the desired result.
	
	Next, since we have
	\begin{align}
	\norm{\vb_{t+1}-\vb_\ast}^2 \le \left(1-\frac{\alpha\sigma_m}{2}+3\sigma^2_M\alpha^2\right)\norm{\vb_t-\vb_\ast}^2 +\left(\frac{\alpha L^2}{\sigma_m}+3L^2\alpha^2\right)\norm{\WB_t-\WB_\ast}^2\norm{\vb_\ast}^2+\left(\frac{2\alpha}{\sigma_m}+3\alpha^2\right)\eta_v^2,
	\end{align}
	therefore, we can get
	\begin{align}
	&\norm{\vb_{t+1}-\vb_\ast}^2 \le \left(1-\frac{\alpha\sigma_m}{2}+3\sigma^2_M\alpha^2\right)\norm{\vb_t-\vb_\ast}^2 \\
	&+\left(\frac{\alpha L^2}{\sigma_m}+3L^2\alpha^2\right)\left[2\left(\frac{1}{{1+\alpha\rho}}\right)^{t}\norm{\WB_0-\WB_\ast}^2+2\eta^2_W(6\alpha +\frac{8}{\rho})^2\right]\norm{\vb_\ast}^2+\left(\frac{2\alpha}{\sigma_m}+3\alpha^2\right)\eta_v^2.
	\end{align}
	Recursively expanding $\norm{\vb_t-\vb_\ast}^2$ yields
	\begin{align}
	\norm{\vb_{t}-\vb_\ast}^2 \le& \left(1-\frac{\alpha\sigma_m}{2}+3\sigma^2_M\alpha^2\right)^t\norm{\vb_0-\vb_\ast}^2 \\&+\left(\frac{2\alpha L^2}{\sigma_m}+6L^2\alpha^2\right)\norm{\WB_t-\WB_\ast}^2\norm{\vb_\ast}^2t\left(1-\frac{\alpha\sigma_m}{2}+3\sigma^2_M\alpha^2\vee \frac{1}{{1+\alpha\rho}}\right)^{t-1}\\&+\left[\left(\frac{2\alpha L^2}{\sigma_m}+6L^2\alpha^2\right)(6\alpha +\frac{8}{\rho})^2\norm{\vb_\ast}^2\eta^2_W+\left(\frac{2\alpha}{\sigma_m}+3\alpha^2\right)\eta_v^2 \right] \frac{1}{\frac{\alpha\sigma_m}{2}-3\sigma^2_M\alpha^2}.
	\end{align}
	Plug in the results from Result~\ref{eigen}, and we have the desired result.
\end{proof}

%
%

\subsubsection*{Proof of Theorem~\ref{training_dynamics}}
\begin{proof}
	First, we prove $\WB_{t+1}\in \WC$. We start by writing
	\begin{align}
	&\Tr(\WB_\ast^\top\WB_{t+1})=\Tr(\WB_\ast^\top\hat\UB)+\Tr(\WB_\ast^\top(\WB_{t+1}-\hat{\UB})) \\
	&\ge \Tr\left(\WB^\top_\ast \frac{\WB_t + \alpha \rho \WB^\ast}{1 + \alpha \rho } \right) -\norm{\WB_\ast^\top}\norm{\WB_{t+1}-\hat{\UB}}\\
	&\ge \frac{\Tr\left(\WB^\top_\ast \WB_t\right)}{1 + \alpha \rho} +\frac{\alpha \rho}{1 + \alpha \rho}-\norm{\WB_\ast^\top}\norm{\WB_{t+1}-\hat{\UB}} \\
	&\ge \frac{\Tr\left(\WB^\top_\ast \WB_t\right)}{1 + \alpha \rho} +\frac{\alpha \rho}{1 + \alpha \rho}-4\alpha\eta_W.
	\end{align}
	Since ${\Tr\left(\WB^\top_\ast \WB_t\right)} \ge {\Tr\left(\WB^\top_\ast \WB_0\right)}/2$ and ${\Tr\left(\WB^\top_\ast \WB_0\right)}\le 1$, we can have
	\begin{align}
	\Tr(\WB_\ast^\top\WB_{t+1}) \ge \frac{\frac{1}{2}}{1+\alpha\rho}{\Tr\left(\WB^\top_\ast \WB_0\right)}+\frac{\alpha \rho}{1 + \alpha \rho}{\Tr\left(\WB^\top_\ast \WB_0\right)} -4\alpha\eta_W.
	\end{align}
	{
		As we have by assumption $4\alpha\eta_W \le \frac{\alpha\rho/2}{1+\alpha\rho}{\Tr\left(\WB^\top_\ast \WB_0\right)}$}, we arrive at
	\begin{align}
	\Tr(\WB_\ast^\top\WB_{t+1}) \ge {\Tr\left(\WB^\top_\ast \WB_0\right)}/2.
	\end{align}
	Therefore, $\WB_{t+1}\in \WC$.

	{Then, we write in a uniform fashion that for both node-level and graph-level tasks that $D=\max \left\lbrace \norm{\vb_0- \vb_\ast}, \sqrt{\frac{\left(\frac{4\alpha L^2}{\sigma_m}+12L^2\alpha^2\right)\norm{\vb_\ast}^2+\frac{2\alpha}{\sigma_m}+3\alpha^2}{\frac{\alpha\sigma_m}{2}-3\sigma_M^2\alpha^2}}\right\rbrace $}, and we
	show $\vb_{t+1}\in \VC$. First, we prove $\norm{\vb_{t+1}-\vb_\ast} \le D$, and it follows directly as
	\begin{align}
	&\norm{\vb_{t+1}-\vb_\ast}^2 \le \left(1-\frac{\alpha\sigma_m}{2}+3\sigma^2_M\alpha^2\right)\norm{\vb_t-\vb_\ast}^2 +\left(\frac{\alpha L^2}{\sigma_m}+3L^2\alpha^2\right)\norm{\WB_t-\WB_\ast}^2\norm{\vb_\ast}^2+\left(\frac{2\alpha}{\sigma_m}+3\alpha^2\right)\eta_v^2\\
	&	\le \left(1-\frac{\alpha\sigma_m}{2}+3\sigma^2_M\alpha^2\right)D^2+\left(\frac{\alpha L^2}{\sigma_m}+3L^2\alpha^2\right)4\norm{\vb_\ast}^2+\left(\frac{2\alpha}{\sigma_m}+3\alpha^2\right) \le D^2.
	\end{align}

	At last, we show $\vb_\ast^\top\vb_{t+1} \ge \rho$. First, we write
	\begin{align}
	\vb_\ast^\top\bar{\gb}^v_t = \vb_\ast^\top\phi_{t,t}(\vb_t-\vb_\ast)+\vb_\ast^\top\left(\phi_{t,t} -\phi_{t,\ast} \right)\vb_\ast
	=\vb_\ast^\top\phi_{t,t}\vb_t-\vb_\ast^\top\phi_{t,\ast} \vb_\ast.
	\end{align}
	Moreover, we obtain
	\begin{align}
	\vb_\ast^\top\bar{\vb}_{t+1} = \vb_\ast^\top({\vb}_{t}-\alpha\bar{\gb}^v_t) &= \vb_\ast^\top\left(\IB-\alpha\phi_{t,t}\right)\vb_t+\alpha\vb_\ast^\top\phi_{t,\ast} \vb_\ast
	&\ge (1-\alpha\sigma_M)\rho + \alpha\sigma_m^\prime\norm{\vb_\ast}^2,
	\end{align}
	{where $\sigma_m^\prime$ is the smallest non-negative eigenvalue of the matrix $\phi_{t,\ast}$, and by definition $\alpha\sigma_m^\prime\norm{\vb_\ast}^2-\alpha\sigma_M\rho \ge \alpha\rho$}, therefore,
	\begin{align}
	\vb_\ast^\top\bar{\vb}_{t+1}\ge \rho+\alpha\rho.
	\end{align}
	
	
	As we can also write
	\begin{align}
	|\vb_\ast^\top\vb_{t+1}-\vb_\ast^\top\bar\vb_{t+1}|\le \alpha\norm{\vb_\ast}\eta_v,
	\end{align}
	{and by the assumption on the number of samples $n$, we have $\norm{\vb_\ast}\eta_v\le\rho$, thus }
	\begin{align}
	\vb_\ast^\top\vb_{t+1} \ge \rho+\alpha\rho-\alpha\rho=\rho.
	\end{align}
\end{proof}
\end{document}